\documentclass[10pt,twocolumn,letterpaper]{article}

\usepackage{wacv}              % To produce the CAMERA-READY version
\usepackage[pdftex]{graphicx}
\usepackage{amsmath}
\usepackage{amssymb}
\usepackage{booktabs}
\usepackage{url}
\usepackage{multirow}
\usepackage{makecell}
\usepackage{booktabs}
\usepackage{nicematrix}
\usepackage{stackrel}
\usepackage{tabularx}
\usepackage{xcolor}
\usepackage{soul}
\usepackage{adjustbox}
\usepackage{float}
\usepackage{pifont}% http://ctan.org/pkg/pifont
\usepackage[accsupp]{axessibility}
\newcommand{\cmark}{\ding{51}}%
\newcommand{\xmark}{\ding{55}}%

\usepackage{colortbl}
\newcommand\boldred[1]{\textcolor{red}{\textbf{#1}}}
\newcommand\boldblue[1]{\textcolor{blue}{\textbf{#1}}}

\newcolumntype{s}{>{\centering\arraybackslash\hsize=.4\hsize}X}
\newcolumntype{a}{>{\centering\arraybackslash\hsize=.15\hsize}X}
\newcolumntype{y}{>{\centering\arraybackslash\hsize=.6\hsize}X}

\definecolor{encoder-color}{RGB}{255,255,255}
\definecolor{decoder-color}{RGB}{255,255,255}
\definecolor{ega-color}{RGB}
{255,255,255}
\newcommand{\hlc}[2][yellow]{{%
    \colorlet{foo}{#1}%
    \sethlcolor{foo}\hl{#2}}%
}

\usepackage[pagebackref,breaklinks,colorlinks]{hyperref}
\usepackage{array}
\usepackage{amssymb,amsmath,amsthm,enumitem}
\usepackage[capitalize]{cleveref}
\crefname{section}{Sec.}{Secs.}
\Crefname{section}{Section}{Sections}
\Crefname{table}{Table}{Tables}
\crefname{table}{Tab.}{Tabs.}

\def\wacvPaperID{2403} % *** Enter the WACV Paper ID here
\def\confName{WACV}
\def\confYear{2024}

\begin{document}

\title{MEGANet: Multi-Scale Edge-Guided Attention Network\\ for Weak Boundary Polyp Segmentation}

\author{Nhat-Tan Bui$^{1}$, Dinh-Hieu Hoang$^{2, 3}$, Quang-Thuc Nguyen$^{2, 3}$, Minh-Triet Tran$^{2, 3}$, Ngan Le$^{1}$\\
$^1$AICV Lab, University of Arkansas, Fayetteville, Arkansas, USA \\
$^2$University of Science, and John von Neumann Institute, VNU-HCM  \\
$^3$Vietnam National University, Ho Chi Minh City, Vietnam \\
}

\maketitle

%%%%%%%%% ABSTRACT
\begin{abstract}
   Efficient polyp segmentation in healthcare plays a critical role in enabling early diagnosis of colorectal cancer. However, the segmentation of polyps presents numerous challenges, including the intricate distribution of backgrounds, variations in polyp sizes and shapes, and indistinct boundaries. Defining the boundary between the foreground (i.e. polyp itself) and the background (surrounding tissue) is difficult. To mitigate these challenges, we propose \textbf{M}ulti-Scale \textbf{E}dge-\textbf{G}uided \textbf{A}ttention Network (MEGANet) tailored specifically for polyp segmentation within colonoscopy images. This network draws inspiration from the fusion of a classical edge detection technique with an attention mechanism. By combining these techniques, MEGANet effectively preserves high-frequency information, notably edges and boundaries, which tend to erode as neural networks deepen. MEGANet is designed as an end-to-end framework, encompassing three key modules: an encoder, which is responsible for capturing and abstracting the features from the input image, a decoder, which focuses on salient features, and the Edge-Guided Attention module (EGA) that employs the Laplacian Operator to accentuate polyp boundaries. Extensive experiments, both qualitative and quantitative, on five benchmark datasets, demonstrate that our MEGANet outperforms other existing SOTA methods under six evaluation metrics. Our code is available at \url{https://github.com/UARK-AICV/MEGANet}.
   
\end{abstract}

%%%%%%%%% BODY TEXT
\section{Introduction}
Colorectal cancer (CRC) is a major health concern due to its high prevalence, ranking as the top gastrointestinal cancer and the third most common cancer. It is second in cancer-related mortality, trailing only behind lung cancer in both genders \cite{crc}. Thus, early CRC detection is of utmost importance. Colonoscopy is the gold standard for CRC examination, yet manual detection and localization of polyps in colonoscopic images are labor-intensive, requiring skilled experts. Consequently, accurate computer-aided polyp segmentation is vital for clinicians to evaluate patients.

In the field of Deep Learning (DL), particularly within the context of computer vision where Convolutional Neural Networks (CNNs) have established dominance, encoder-decoder network architectures \cite{unet, unet++, sfa, pranet, msnet, sanet, ho2021point, nguyen20213d, isensee2019no, tran2022ss, dam-al, rivf} have demonstrated significant success in the realm of medical image segmentation. While methods such as U-Net++ \cite{unet++}, SFA \cite{sfa}, PraNet \cite{pranet}, MSNet \cite{msnet}, and SANet \cite{sanet} are mainly designed for polyp segmentation, the accuracy of the segmentation heavily hinges on the amalgamation of encoded feature maps from various scales in the contracting path and the semantically enriched decoded feature maps in the expanding path. Despite notable advancements, these methodologies still grapple with the challenge of preserving high-frequency information, a critical aspect of medical imaging. Particularly, the presence of variable mucous membranes surrounding the polyps, differing in shape, color, and texture, contributes to complex and diverse polyp borders. This complexity, combined with the downscaled encoding, challenges the maintenance of border details and the improvement of segmentation during decoding, resulting in imprecise polyp boundary generation. Our insight underscores that edge information obtained through conventional image processing techniques tends to be more straightforward and precise than edges extracted by CNN-based methods, especially when training data is scarce. Hence, a promising strategy involves revisiting classical image processing-based edge extraction techniques. This approach holds the potential for addressing the issue of weak boundaries in medical imaging. In our present study, we harness the capabilities of classical edge features by introducing the Edge-Guided Attention (EGA) module. This module is designed to function across multiple scales, spanning from low-level to high-level features. Its primary objective is to compel the model to focus on edge-related information, thereby enhancing predictions at each decoder level. Importantly, the EGA module achieves this objective without succumbing to noise or encountering the semantic gap. Our EGA approach capitalizes on classical edge extraction methods to augment the accuracy of medical image segmentation. The EGA module, operating at multiple scales and targeting edge-related features, addresses the challenge of weak boundaries in a noise-resistant manner, enriching the segmentation predictions across the decoder's levels.

As a result, this paper introduces the \textbf{M}ulti-Scale \textbf{E}dge-\textbf{G}uided \textbf{A}ttention Network (MEGANet), a novel and innovative approach that integrates an EGA module into the U-Net architecture during the decoding process. The primary goal of MEGANet is to preserve crucial edge and boundary information effectively. In essence, MEGANet comprises three key modules: (i) Encoder, responsible for capturing the visual representation of the input image, akin to the encoder in the U-Net architecture. (ii) Decoder aims to extract salient features, following similar settings as in the U-Net decoder architecture. (iii) EGA module, this distinctive module leverages the Laplacian operator to preserve high-frequency information, particularly edges. The EGA module operates on both the embedding feature and multi-level predictions. This strategic combination empowers the model to accentuate intact edge details and polyp boundaries across various scales. In summary, this paper's main contributions are: (i) We explore the potential of the Laplacian operator, a parameter-free method, to enhance the segmentation of weak boundary objects like polyps by preserving high-frequency edge information. (ii) We present a novel architecture, MEGANet, that addresses the challenge of supplementing low-level boundary information using the Laplacian operator. (iii) We extensively evaluate our method on five benchmark datasets, i.e., Kvasir-SEG \cite{kvasir}, CVC-ClinicDB \cite{clinicdb}, CVC-ColonDB \cite{colondb}, ETIS \cite{etis}, and EndoScene \cite{cvc300}, to demonstrate its effectiveness.

\section{Related Work}
\label{sec:RelatedWork}

The widely adopted U-Net \cite{unet} architecture, known for its effectiveness in medical image segmentation, has been applied to polyp segmentation. U-Net++ \cite{unet++}, an enhanced variant, addresses semantic gaps through nested skip connections. Although these concepts have broader applications beyond medical imaging, they predominantly focus on enhancing feature learning rather than medical-specific challenges.

Various approaches heavily leverage boundary information in addressing the challenge of weak boundaries in medical imaging. SFA \cite{sfa} introduces an extra decoder for boundary prediction and employs a boundary-sensitive loss to exploit area-boundary relationships. PraNet \cite{pranet} uses parallel partial decoders and reverse attention modules to progressively extend object regions by incorporating edge features. ACSNet \cite{acsnet} combines local and global context to accommodate varying polyp sizes. NB-AC \cite{le2021narrow} introduces narrow band active contour attention when considering weak boundary is a confusing case that needs more attention. \cite{le2021offset} presents offset curve loss to give more attention to the boundary. DAM-AL \cite{dam-al} employs dilated attention for long-range relationships and introduces a novel loss mechanism. MSNet \cite{msnet} introduces the multi-scale subtraction module for mitigating redundant and obtaining complementary information. PEFNet \cite{pefnet} focuses on the positional information of the polyp objects in the skip connection with the EfficientNetV2 \cite{efficientnetv2} encoder. M$^2$UNet \cite{m2unet} integrates MetaFormer \cite{metaformer} and multi-scale information for enhanced context exploitation.

In contrast, our approach explicitly addresses the weak boundary issue by incorporating high-frequency edge information obtained through typical image processing techniques. A fundamental limitation of prior methods is their inability to accurately reconstruct input image edges, as CNN-based features are not optimized for this purpose. We opt for the Laplacian operator to extract and retain high-frequency features, particularly edge details, in polyp images. Laplacian, a second-order derivative operator, yields more meaningful edge structures than hand-crafted first-order derivative methods like Sobel \cite{sobel} or Prewitt \cite{prewitt} without adding computational complexity. Laplacian has been successfully applied in many image processing problems, such as style transfer \cite{laplaciansteered, icmv}, image super-resolution \cite{resolution},  image synthesis \cite{imagesys}, image deraining \cite{derain}, image-to-image translation \cite{imagetoimage}, etc.

\setlength{\abovedisplayskip}{2pt}
\setlength{\belowdisplayskip}{2pt}

\section{Proposed MEGANet}
\label{sec:ProposedMethod}

\begin{figure*}[t]
\begin{center}
\includegraphics[width=0.92\textwidth]{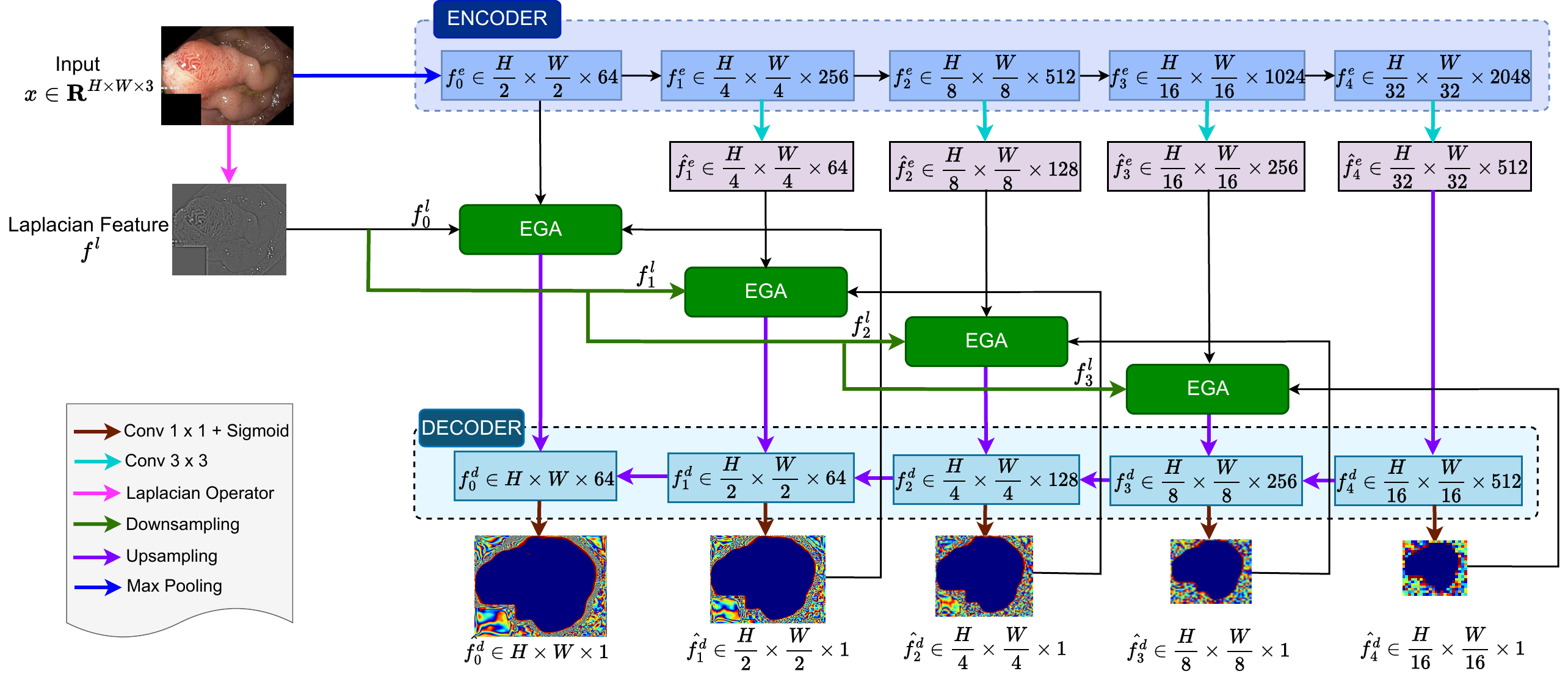}
\caption{Overall architecture of proposed MEGANet(Res2Net-50), including three modules, i.e., \emph{\hlc[encoder-color]{encoder}} and \emph{\hlc[decoder-color]{decoder}}, which utilizes a U-Net to extract visual representations, and the novel \emph{edge attention} module, denoted as the \hlc[ega-color]{Edge-Guided Attention} (\hlc[ega-color]{EGA}) module, designed to retain high-frequency details effectively. In this context, $H, W$ represents the input height and width.} 
\vspace{-0.2in}
\label{fig:EGANet}
\end{center}
\end{figure*}

Our proposed MEGANet architecture comprises three main modules: an encoder, a decoder, and an EGA (Edge-Guided Attention) module, as depicted in Figure \ref{fig:EGANet}. The encoder, located in the contracting path, captures context and high-level features from the input polyp image. It encompasses five convolutional blocks and results in encoding feature $f^e_i$ at the $i^{th}$ layer. 
%The first layer of the contracting path is structured with a combination of operations: Convolution, Batch Normalization, Rectified Linear Unit (ReLU), and Maxpooling. Subsequent layers in the architecture replace Maxpooling with Convolution, Stride 2. 
On the other hand, the decoder, situated in the expanding path, leverages the high-level features acquired by the encoder to generate decoding maps $f^d_i$ at the $i^{th}$ layer that matches the original resolution of the input image. To showcase the effectiveness of our MEGANet as well as to conduct a fair comparison with existing work, we evaluate its performance using two distinct backbone networks: ResNet-34 \cite{resnet} and Res2Net-50 \cite{res2net}. 

In the expanding path, pooling and strided convolution layers are employed to downsample feature maps, reducing the volume of information for processing. While downsampling layers offer significant advantages for constructing deep architectures, it's noteworthy that conventional CNNs often suffer from the loss of information as downsampling layers accumulate at deeper levels. Recognizing the importance of preserving such critical information in medical segmentation, we introduce a novel Edge-Guided Attention module (EGA) that operates between the two aforementioned paths at every resolution level. 
In MEGANet, the output of the contracting path serves as an input to the EGA module, and subsequently, the output of the EGA module feeds into the expanding path. This establishes a coherent linkage between the contracting and expanding paths through the intermediary EGA modules. The subsequent subsection will delve into a comprehensive explanation of the EGA module's functioning and attributes. We detail the EGA module in the following section.

\subsection{Edge-Guided Attention Module (EGA)}
The primary objective of the EGA module is to robustly preserve edge information across multiple scales, effectively addressing the issue of weak boundaries in polyp segmentation. Additionally, the EGA module plays a pivotal role in bridging the semantic gap between the low-level features extracted by the encoder and the high-level features produced by the decoder prior to their fusion. 

\begin{figure*}[t]
\centering
\includegraphics[width=0.94\textwidth]{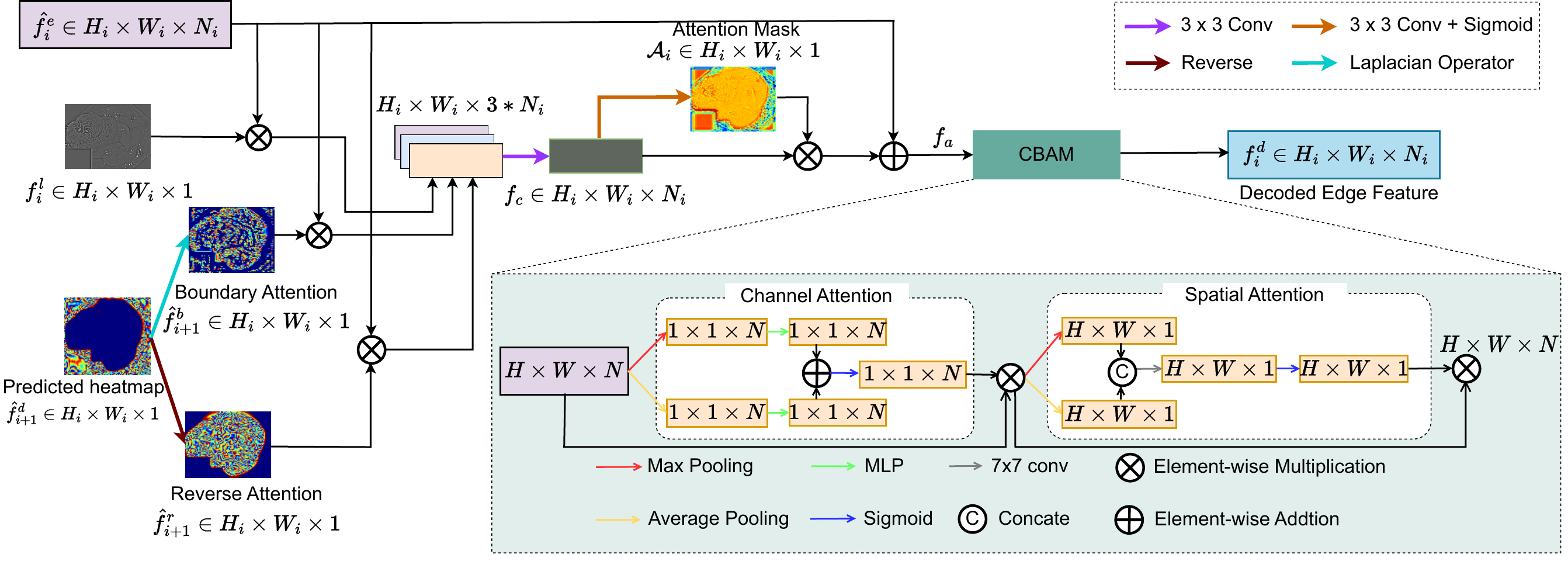}
\caption{The architecture of our EGA block, which takes embedding feature $\hat{f}_i^e \in H_i\times W_i \times N_i$, edge information by Laplacian feature $f_i^l \in H_i\times W_i \times 1$ and higher-level predicted feature $\hat{f}_{i+1}^d \in H_i\times W_i \times 1$ as its input. $H_i$, $W_i$, $N_i$ denote height, width, and the number of channels of the input at the layer $i^{th}$. } 
\vspace{-0.2in}
\label{fig:EGA-module}
\end{figure*}

At the $i$-th layer, the EGA module takes three inputs: the embedding feature $\hat{f}_{i}^{e}$ from the encoder, the high-frequency feature $f_{i}^{l}$ obtained through classical image processing methods, i.e., edge detector, and the higher-level predicted feature $\hat{f}_{i+1}^{d}$ generated by the decoder. The EGA module processes these inputs and generates an output feature map denoted as $f_{i}^{d}$. The detailed process by which the inputs are processed and the specific operations performed by the EGA module are elaborated below. Refer to Figure \ref{fig:EGA-module} for a visual representation of the EGA module.

\subsubsection{EGA Input}
The EGA module takes three inputs: encoded visual feature $f^e$ from the contracting path, decoded predicted feature $\hat{f}^d$ from a higher layer in the expanding path, and high-frequency feature $f^l$ from the Laplacian operator. 

\noindent
\textbf{\textit{Encoded visual feature.}} Each resultant encoding feature map $f^e_i$ from the contracting path then undergoes a $3\times 3$ convolutional operation to reduce the number of channels, producing a distinctive encoded visual feature denoted as $\hat{f}^e_i\in \mathbb{R}^{H_i \times W_i \times N_i}$ at the $i^{th}$ layer. However, we use $f^e_0$ for the first EGA module as its number of channels is already small. For convenience, we still write $\hat{f}^e_0$ instead of $f^e_0$.

\noindent
\textbf{\textit{Decoded predicted feature.}} The second input of the EGA at the $i^{th}$ layer is the decoded predicted feature from a higher layer, specifically at $(i+1)^{th}$ layer, denoted as $\hat{f}^d_{i+1} \in \mathbb{R}^{H_i \times W_i \times 1}$. This decoded predicted feature, $\hat{f}^d_{i+1}$, is derived from the decoding feature $f^d_{i+1}$, which is the output of the EGA at the $(i+1)^{th}$ layer.

\noindent
\textbf{\textit{High-frequency feature.}}
To validate the effectiveness of EGA, we opt for the Laplacian pyramid method, an efficient technique for preserving high-frequency details, namely, image edge information. It is important to note that the Laplacian operator is a second-order derivative operator primarily employed for edge detection. Nevertheless, due to its susceptibility to noise, practical application entails an initial smoothing of the original image using a Gaussian filter. This modified process is known as the Laplacian of Gaussian (LoG). To enhance computational efficiency, the LoG method is itself approximated using the Difference of Gaussian (DoG) operator. This operator essentially functions as a highpass filter, proficiently retaining the most salient high-frequency attributes within the image. The Laplacian pyramid can be considered a sequence of cascading approximations of the DoG operator. In essence, the Laplacian pyramid encapsulates crucial low-level details across various scales.
\begin{equation}
\begin{split}
I_{k} & = I, \text{if } k=0. \\
I_{k} & = d(g(I_{k-1})), \text{if } k\geq1.
\end{split}
\label{eq:1}
\end{equation}
where $I$ is the input image, $g$ is the convolution operator with Gaussian filter, and $d$ denotes the 2 $\times$ downsampling operation, respectively. Each level $L_k$ of the Laplacian pyramid is attained from the Gaussian pyramid by subtracting from the current level $I_k$ the upsampled version ($u$) of the smaller one $I_{k+1}$.
\begin{equation}
    L_k = I_k - u(I_{k+1}).
\label{eq:2}
\end{equation}

The Laplacian operator captures second-order variations within the input image, rendering it a parameter-free and harmonious approach to extracting high-frequency details, including edges, contours, and more. These high-frequency features play a pivotal role in discerning the unique characteristics of the polyp, particularly in distinguishing it from the adjacent mucous membrane. As a result, the high-frequency feature at $i^{th}$ level $f^l_i \in \mathbb{R}^{H_i \times W_i \times 1}$ corresponding to the level-1 image in the Laplacian pyramid is supplied to the $i^{th}$ EGA module. In general, we denote $f^l = L_1(I)$. 

This particular pyramid level is selected because, as resolution diminishes, the finer edge details undergo significant degradation owing to the repeated application of the Gaussian filter. Kindly refer to the supplementary material provided for a visual representation of the level-1 Laplacian pyramid. It's important to note that our selection of the Laplacian operator and its associated techniques serves as a proof of concept for our approach, albeit without an exhaustive practical evaluation. The decision to employ the Laplacian pyramid as the mechanism for extracting high-frequency information is grounded in its simplicity, minimal computational overhead, and the quality of information it retains. It's worth mentioning that the model's performance could potentially be enhanced through an exhaustive search for an appropriate edge detection technique, as the choice of such a technique can significantly impact the overall results.

\subsubsection{EGA Procedure}

At the $i^{th}$ layer, the EGA module integrates multiple components, including the encoded visual feature $\hat{f}^e_i\in \mathbb{R}^{H_i \times W_i \times N_i}$, the predicted feature map at a higher layer $(i+1)^{th}$, represented as $\hat{f}^d_{i+1} \in \mathbb{R}^{H_i \times W_i \times 1}$ and the high-frequency feature $f^l_i \in \mathbb{R}^{H_i \times W_i \times 1}$. Generally, a Laplacian pyramid consists of high-frequency components at multiple scales. However, the high-frequency component at the $i^{th}$ level of the Laplacian pyramid is obtained by performing Gaussian filtering, followed by a 2 $\times$ downsampling of level $(i-1)^{th}$. This process can reduce the magnitude of the high-frequency information at the $i^{th}$ level. To preserve the high-frequency information $f_i^l$ at every level, we propose to derive $f_i^l$ from the base level $f_0^l$, which retains the high-frequency information most effectively. The formulation for calculating $f_i^l$ is presented in Equation \ref{eq:fl}.
\begin{equation}
\begin{split}
    f_0^l & = f^l, \text{ where } f^l = L_1(I). \\
    %f^l_i & = d^i(f^l).\\
    f^l_i & = (d(f^l))^{i}, \text{ if } i \geq 1.
\end{split}   
\label{eq:fl}
\end{equation}
where $d$ is 2 $\times$ downsamling and  $(d(f^l))^{i}$ is $i$ times 2 $\times$ downsamling, i.e., $d(d(...d(f^l)))$. 

Drawing upon insights from \cite{reverseattention}, we apply a decomposition method to the higher-level predicted map $\hat{f}^d_{i+1}$, generating two distinct attention maps: i) a reverse attention map $\hat{f}^r_{i+1}$, calculated as $\hat{f}^r_{i+1} = 1 - \hat{f}^d_{i+1}$ to re-evaluate and refine the imprecise prediction map from the higher layer, and ii) a boundary attention map $\hat{f}^b_{i+1}$, derived by applying the Laplacian operator, i.e., $\hat{f}^b_{i+1} = L_{0}(\hat{f}^d_{i+1})$. Subsequently, we execute element-wise multiplication between the three attention maps, namely ${f}^l_{i}$, $\hat{f}^b_{i+1}$, and $\hat{f}^r_{i+1}$, with the current encoder features $\hat{f}^e_{i}$. This process culminates in the creation of a combined feature $f_c$, characterized as follows:

\begin{equation}
    f^c_i = \text{Conv}([({f}^l_{i} \otimes \hat{f}^e_i), (\hat{f}^b_{i+1} \otimes \hat{f}^e_i), (\hat{f}^r_{i+1} \otimes \hat{f}^e_i)])
\end{equation}
where $[.]$ denotes concatenation. Recognizing that edge information could potentially encompass noise and superfluous details unhelpful for polyp segmentation; we introduce an attention mask denoted as $\mathcal{A}_i$ at level $i$. This mask serves the purpose of guiding the model's attention toward vital regions while simultaneously suppressing background noise and redundant information. The attention feature map at the $i^{th}$ layer, $f^a_i$, is defined as follows:

\begin{align}
f^a_i = \hat{f}^e_i + (f^c_i \otimes \mathcal{A}_i), \text{where } \mathcal{A}_i = \sigma(\text{Conv}(f^c_i))
\end{align}

\begin{figure*}[!h]
\setkeys{Gin}{width=\linewidth,height=\linewidth}
\renewcommand{\arraystretch}{0.4}
\begin{center}
\setlength{\tabcolsep}{2pt}
\begin{tabularx}{0.9\textwidth}{XXXXXX}  
    \centering Image & \centering $\mathcal{A}_3$ & \centering $\mathcal{A}_2$ & \centering $\mathcal{A}_1$ & \centering $\mathcal{A}_0$ & \centering GT \endline
    
    \includegraphics{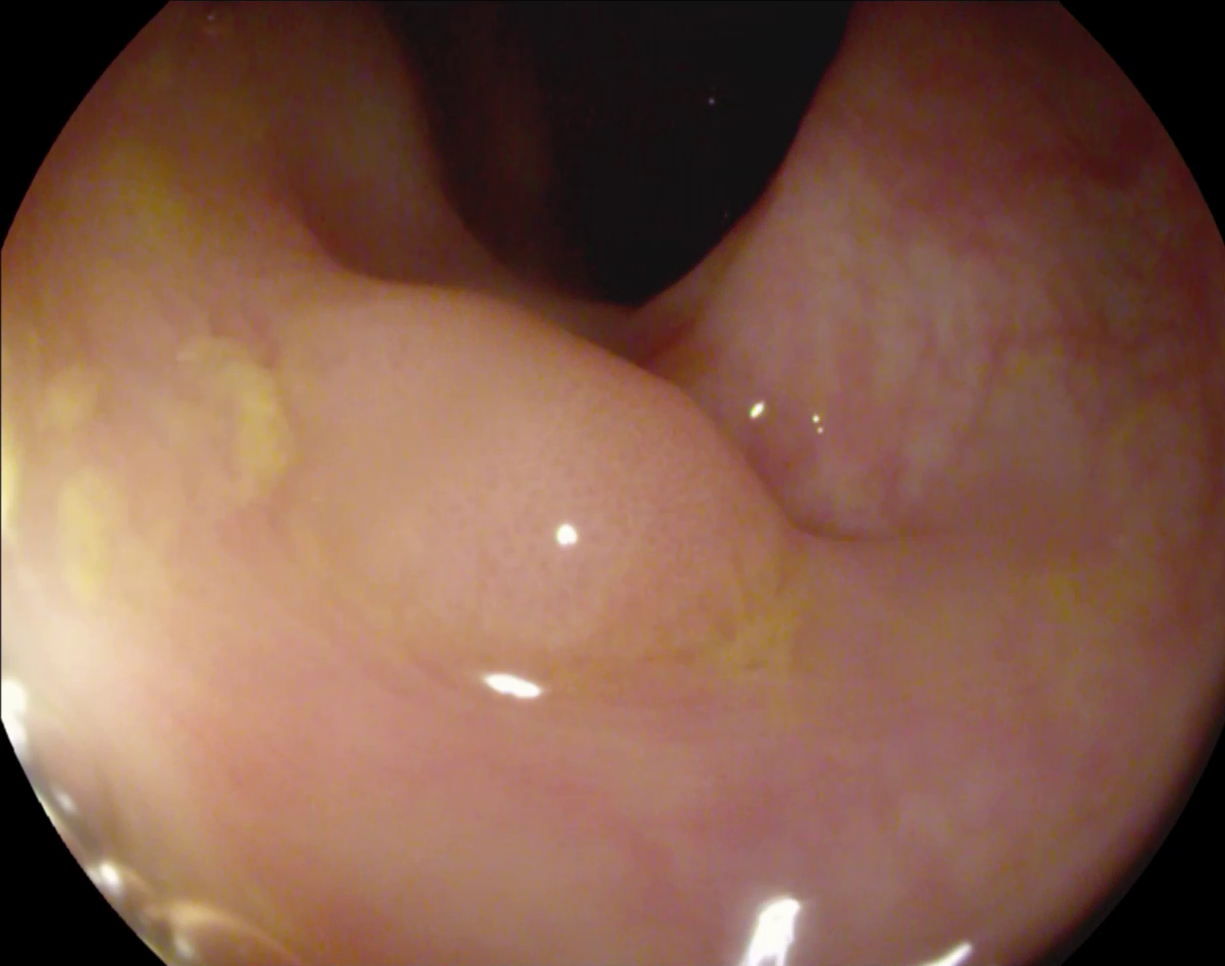} & \includegraphics{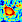} & 
    \includegraphics{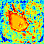} & 
    \includegraphics{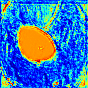} & 
    \includegraphics{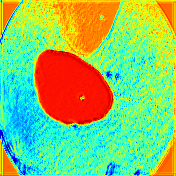} & 
    \includegraphics{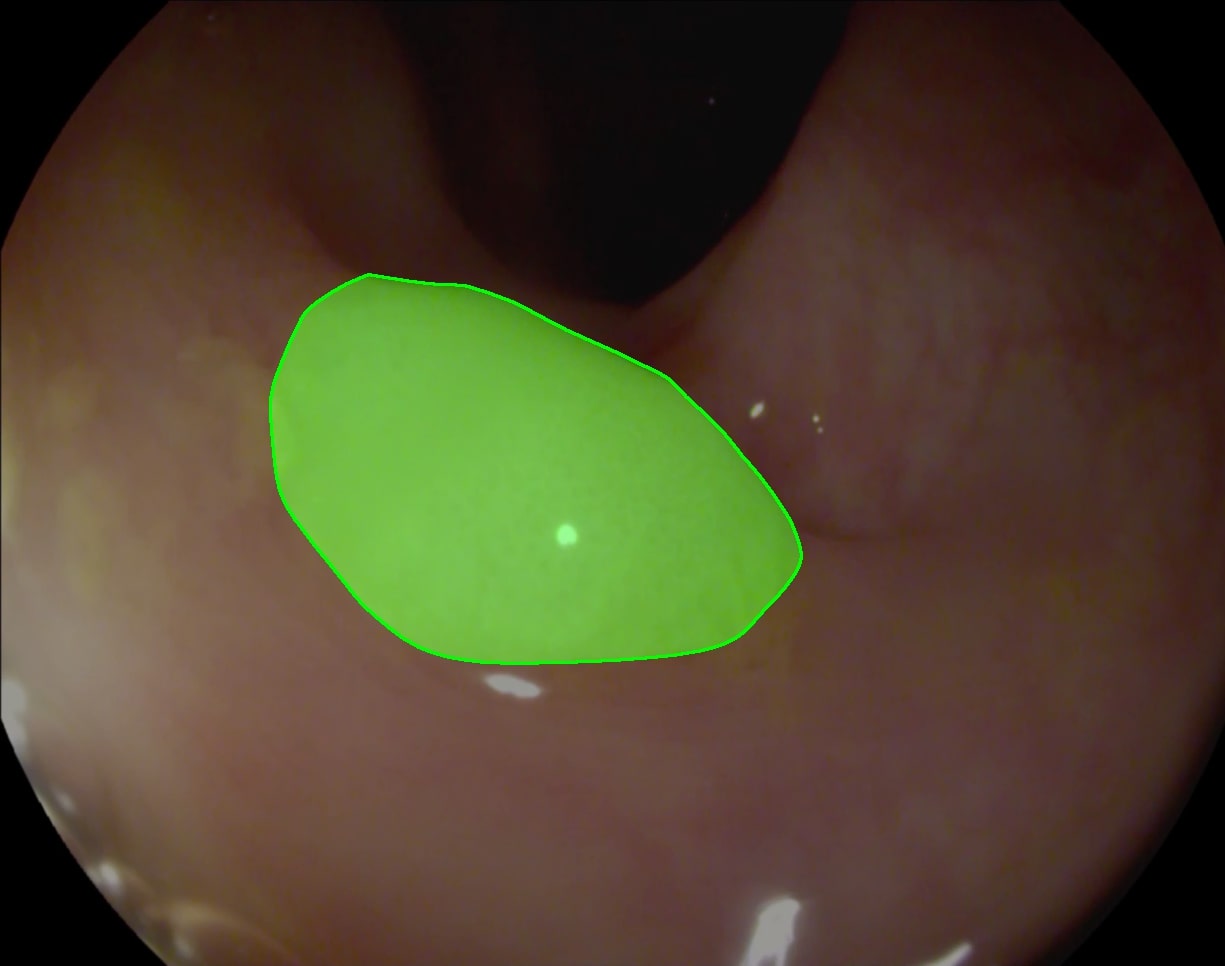} \\

    \includegraphics{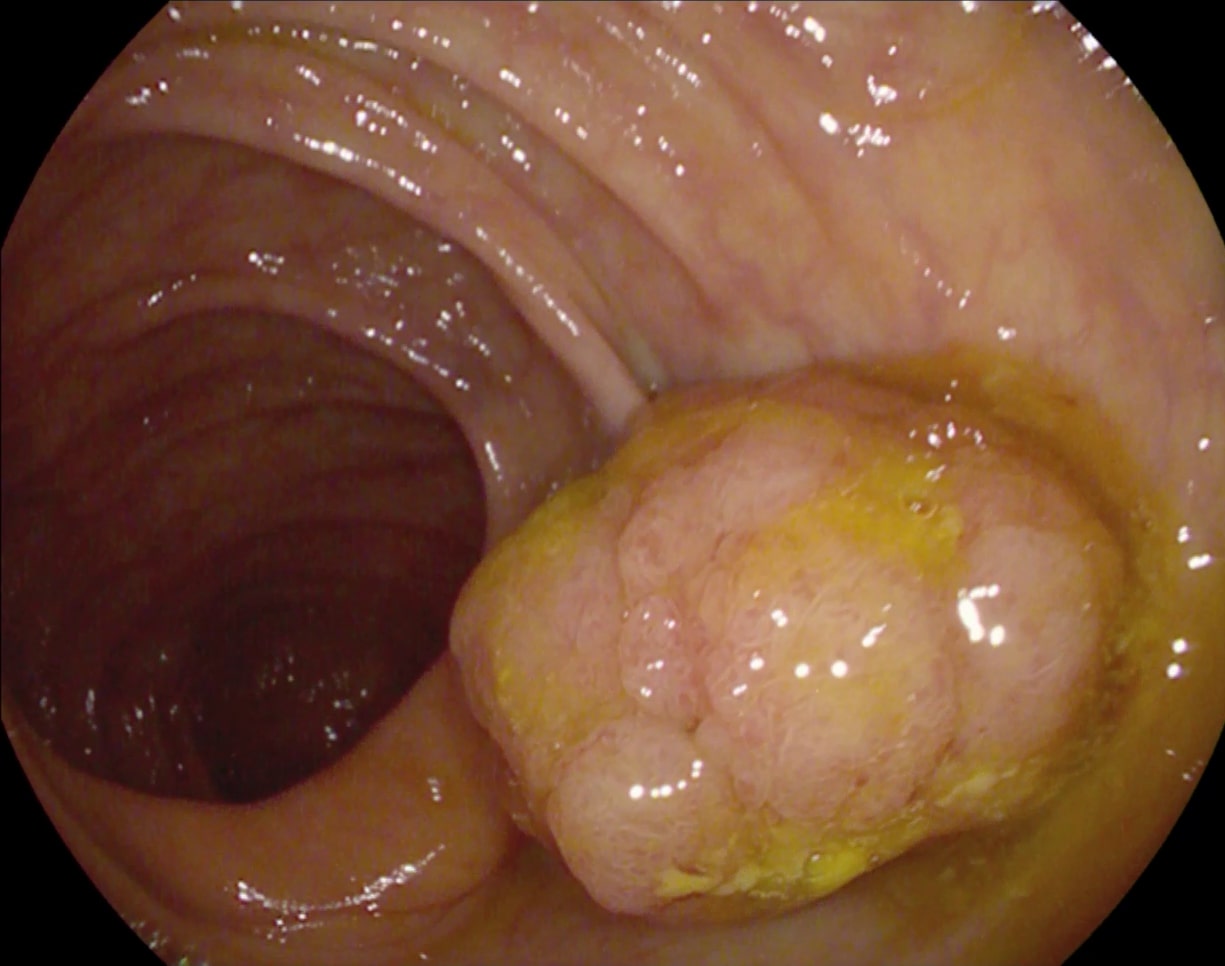} & \includegraphics{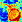} & 
    \includegraphics{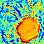} & 
    \includegraphics{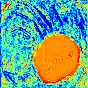} & 
    \includegraphics{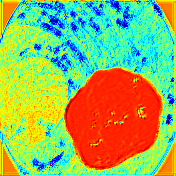} & 
    \includegraphics{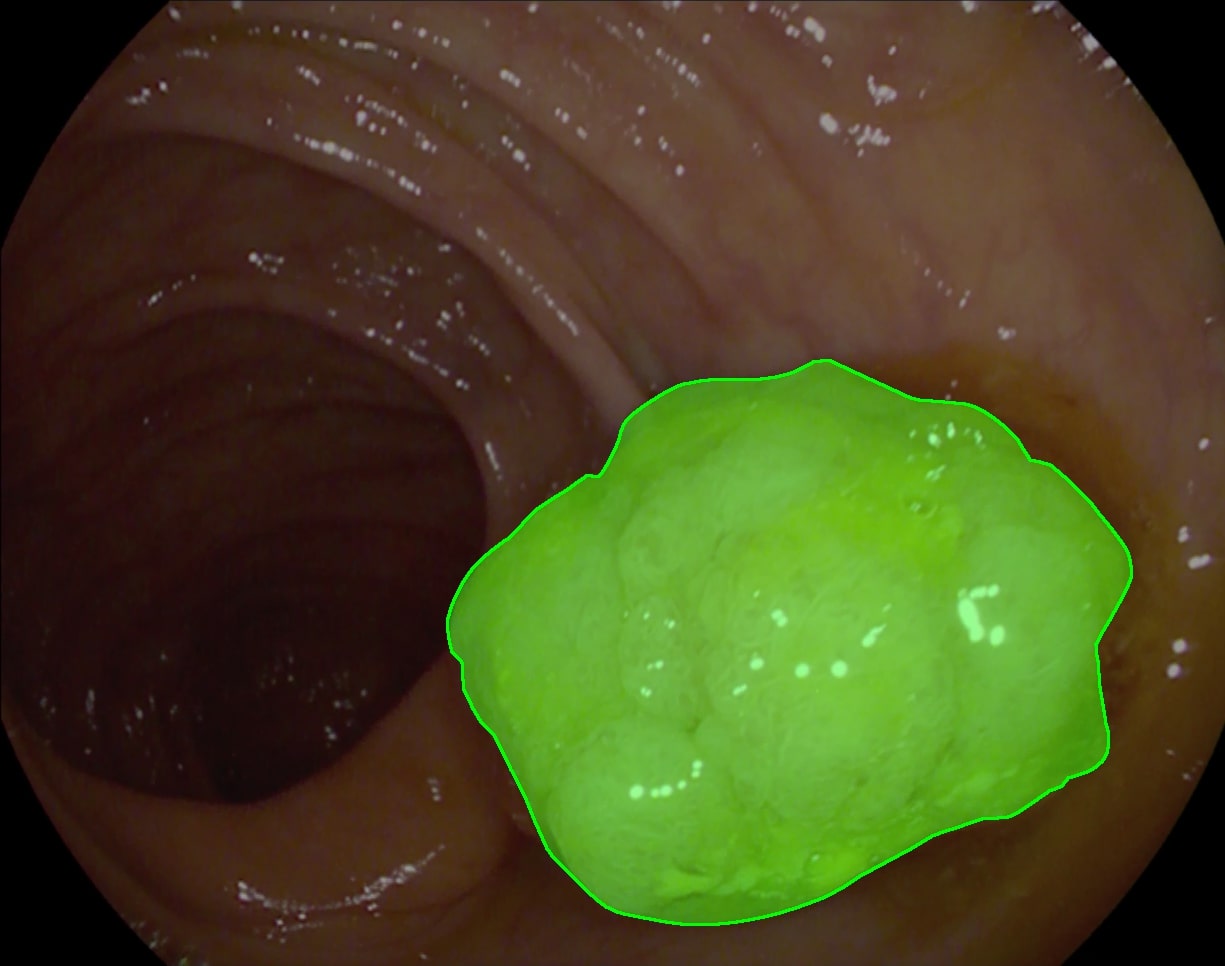} \\
\end{tabularx}
\vspace{-0.01in}
\caption{Heatmap visualization of attention mask $\mathcal{A}_i$ in EGA module at different $i^{th}$ layer.} 
\vspace{-0.2in}
\label{heatmap}
\end{center}
\end{figure*}

In this context, the symbol $\sigma$ signifies the sigmoid function. As depicted in Figure \ref{heatmap}, the attention masks $\mathcal{A}_i$ exhibit markedly elevated values precisely at pixels located along the edges of the polyp. In simpler terms, the fusion of deep features and Laplacian features empowers the model to prioritize the polyp's edge accurately. Building upon insights from \cite{cbam}, we subject the attention feature $f^a_i$ to a $\mathtt{CBAM}$ (Convolutional Block Attention Module) for recalibration purposes. This step facilitates the capture of feature correlations between the boundary and the background region. The $\mathtt{CBAM}$ comprises two consecutive blocks: channel attention, concentrating on the channel dimension, and spatial attention, centering on the spatial dimension. In the channel attention, the attention feature map $f^a_i$ is refined by convolution with kernels $1 \times 1 \times N_i$. In the spatial attention, spatial kernels of  $H_i \times W_i \times 1$ are used. The configuration of this module is visually depicted in Figure \ref{fig:EGA-module}. This figure uses $H, W, N$ for a general case. Consequently, this process yields a refined decoding feature $f^d_i$, i.e.,  $f^d_i = \mathtt{CBAM}(f^a_i)$.

% \begin{figure*}[t]
% \centering
% \includegraphics[width=0.94\textwidth]{Tex/Heatmap-EGA/att.jpeg}
% \caption{Heatmap visualization of attention mask $\mathcal{A}_i$ in EGA module at different $i^{th}$ layer.} 
% \vspace{-0.2in}
% \label{heatmap}
% \end{figure*}

\subsection{Objective function}
We employ a combination of the binary cross-entropy loss ($\mathcal{L}_{BCE}$) and the dice loss ($\mathcal{L}_{Dice}$) as our network's objective functions for training, taking into account their established efficacy as demonstrated in \cite{acsnet}. Consequently, the objective function for our MEGANet can be formally defined as follows: 
\begin{equation}
    \mathcal{L}_{EGA} = \sum_{i=1}^{D}{ \mathcal{L}_{BCE}(\hat{f}^d_{i}, f_i) + \mathcal{L}_{Dice}(\hat{f}^d_{i}, f_i)}
\end{equation}
where $D$ is the number of decoder layers. We maintain the standard setting of 5 as inherited from U-Net. $\hat{f}^d_i$ is predicted feature map at the $i^{th}$ decoding layer and $f_i$ is groundtruth polyp segmentation at scale $i^{th}$. 

\section{Experiments}
\label{sec:Experiments}

\subsection{Datasets and Evaluation Metrics}

\noindent
\textbf{Datasets.}
We evaluate our proposed MEGANet on five standard benchmark datasets: Kvasir-SEG \cite{kvasir}, CVC-ClinicDB \cite{clinicdb}, CVC-ColonDB \cite{colondb}, ETIS \cite{etis} and EndoScene \cite{cvc300}. Note that the EndoScene \cite{cvc300} composes 912 images of two subsets, CVC-ClinicDB and CVC-300.

To conduct a fair comparison, we follow the same experiment setup in \cite{pranet}, which selects 1,450 images from Kvasir (900 images), and CVC-ClinicDB (550 images) for the training set while 798 images from Kvasir (100 images), CVC-ClinicDB (62 images), CVC-ColonDB (380 images), ETIS (196 images), and CVC-300 (60 images) for testing. This setting is challenging since the evaluation procedure is conducted across the different datasets with a wide range of resolutions (720 x 576 up to 1,920 x 1,072 in Kvasir, 384 x 288 in CVC-ClinicDB) and varied image-acquiring processes, which introduce high variance across these datasets in the size and shape of the polyps.

\noindent
\textbf{Evaluation Metrics.}
% \hl{Please list all definitions of metrics with predicted $\hat{\textbf{y}}$ and groundtruth $\textbf{y}$}
To conduct a comprehensive evaluation and comparisons with other methodologies, we follow existing SOTA approaches, employing five different metrics, i.e., mean Dice (mDice), mean IoU (mIoU), the weighted F-measure ($F_{\beta}^{w}$) \cite{fmeasure}, the structure measure ($S_{\alpha}$) \cite{smeasure}, the enhanced-alignment measure ($E_{\phi}^{max}$) \cite{emeasure} and mean absolute error (MAE). These metrics serve the dual purpose of assessing the performance of our method in relation to ground truth labels, i.e., between prediction $\hat{f}^d$ and ground truth $f$, as well as facilitating comparative analysis with other existing techniques. The effectiveness of those metrics in polyp segmentation is discussed in \cite{pranet, msnet}.

\setlength{\tabcolsep}{5pt}

\begin{table*}[t]
\caption{Performance comparison between our MEGANet and other existing SOTA methods on Kvasir, CVC-300 (EndoScene), ColonDB, and ETIS datasets. The highest and second highest scores are shown in \textbf{bold} and \underline{underline}, respectively. All metrics are in (\%).}\label{tab1}
\resizebox{\textwidth}{!}{
\begin{NiceTabular}{c|l|*{6}{c}|*{6}{c}}
    \toprule
    & \multirow{2}{*}{Methods} & \multicolumn{6}{c}{Kvasir-SEG (seen)}  & \multicolumn{6}{c}{CVC-300 (EndoScene) (unseen)}\\
    \cmidrule(lr){3-8}\cmidrule(lr){9-14}
    & &  mDice   $\uparrow$ & mIoU   $\uparrow$ & $F_{\beta}^{w}$   $\uparrow$ & $S_{\alpha}$   $\uparrow$ &  $E_{\phi}^{max}$   $\uparrow$ & MAE   $\downarrow$
    & mDice  $\uparrow$ & mIoU   $\uparrow$ & $F_{\beta}^{w}$ $\uparrow$ & $S_{\alpha}$   $\uparrow$ &  $E_{\phi}^{max}$   $\uparrow$ & MAE   $\downarrow$\\
    \hline
    
    \multirow{8}{*}{\rotatebox[]{90}{SOTA methods}}& U-Net\cite{unet} & 81.8 & 74.6 & 79.4 & 85.8 & 89.3 & 5.5 & 71.0 & 62.7 & 68.4 & 84.3 & 87.6 & 2.2 \\
    & U-Net++\cite{unet++} & 82.1 & 74.3 & 80.8 & 86.2 & 91.0 & 4.8 & 70.7 & 62.4 & 68.7 & 83.9 & 89.8 & 1.8\\
    & SFA\cite{sfa} & 72.3 & 61.1 & 67.0 & 78.2 & 84.9 & 7.5 & 46.7 & 32.9 & 34.1 & 64.0 & 81.7 & 6.5\\
    & PraNet\cite{pranet} & 89.8 & 84.0 & 88.5 & 91.5 & 94.8 & 3.0 & 87.1 & 79.7 & 84.3 & 92.5 & \textbf{97.2} & 1.0\\
    & SANet\cite{sanet} & 90.4 & 84.7 & 89.2 & 91.5 & 95.3 & 2.8 & 88.8 & 81.5 & 85.9 & \underline{92.8} & \textbf{97.2} & \underline{0.8} \\ 
    & MSNet\cite{msnet} & 90.7 & \underline{86.2} & 89.3 & \textbf{92.2} & 94.4 & 2.8 & 86.9 & 80.7 & 84.9 & 92.5 & 94.3 & 1.0\\
    % & LDNet(MICCAI'22) & 90.7 & 91.0 & -- & -- & -- & -- & -- & -- & -- & -- & -- & --\\
    & PEFNet\cite{pefnet} & 89.2 & 83.3  & -- & -- & -- & 2.9 & 87.1 & 79.7 & -- & -- & -- & 1.0\\
    & M$^2$UNet\cite{m2unet} & 90.7 & 85.5 & -- & -- & -- & \textbf{2.5} & \underline{89.0} & \underline{81.9} & -- & -- & -- & \textbf{0.7}\\
    
    \hline
    \multicolumn{2}{l}{\textbf{MEGANet(ResNet-34)}}& \underline{91.1} & 85.9 & \underline{90.4} & 91.6 & \underline{95.4} & \underline{2.6} & 88.7 & 81.8 & \underline{86.3} & 92.4 & 95.9 & 0.9\\
    \multicolumn{2}{l}{\textbf{MEGANet(Res2Net-50)}} &  \textbf{91.3} & \textbf{86.3} & \textbf{90.7} & \underline{91.8} & \textbf{95.9} & \textbf{2.5} & \textbf{89.9} & \textbf{83.4} & \textbf{88.2} & \textbf{93.5} & \underline{96.9} & \textbf{0.7}\\

\hline \hline
    & \multirow{2}{*}{} & \multicolumn{6}{c}{ColonDB (unseen)}  & \multicolumn{6}{c}{ETIS (unseen)}\\
    \cmidrule(lr){3-8}\cmidrule(lr){9-14}
    & & mDice $\uparrow$ & mIoU $\uparrow$ & $F_{\beta}^{w}$ $\uparrow$ & $S_{\alpha}$ $\uparrow$ &  $E_{\phi}^{max}$ $\uparrow$ & MAE $\downarrow$
    & mDice $\uparrow$ & mIoU $\uparrow$ & $F_{\beta}^{w}$ $\uparrow$ & $S_{\alpha}$ $\uparrow$ &  $E_{\phi}^{max}$ $\uparrow$ & MAE $\downarrow$\\
    \hline
    
    \multirow{8}{*}{\rotatebox[]{90}{SOTA methods}} & U-Net\cite{unet} & 51.2 & 44.4 & 49.8 & 71.2 & 77.6 & 6.1 & 39.8 & 33.5 & 36.6 & 68.4 & 74.0 & 3.6 \\
    & U-Net++\cite{unet++} & 48.3 & 41.0 & 46.7 & 69.1 & 76.0 & 6.4 & 40.1 & 34.4 & 39.0 & 68.3 & 77.6 & 3.5 \\
    & SFA\cite{sfa} & 46.9 & 34.7 & 37.9 & 63.4 & 76.5 & 9.4 & 29.7 & 21.7 & 23.1 & 55.7 & 63.3 & 10.9 \\
    & PraNet\cite{pranet} & 70.9 & 64.0 & 69.6 & 81.9 & 86.9 & 4.5 & 62.8 & 56.7 & 60.0 & 79.4 & 84.1 & 3.1 \\
    & SANet\cite{sanet} & 75.3 & 67.0 & 72.6 & 83.7 & 87.8 & 4.3 & \underline{75.0} & 65.4 & 68.5 & \underline{84.9} & \underline{89.7} & \textbf{1.5} \\
    & MSNet\cite{msnet} & 75.5 & 67.8 & 73.7 & 83.6 & 88.3 & 4.1 & 71.9 & 66.4 & 67.8 & 84.0 & 83.0 & 2.0 \\
    % & LDNet(MICCAI'22) & \underline{78.4} & \textbf{83.4} & -- & -- & -- & -- & \underline{74.4} & \textbf{82.3} & -- & -- & -- & --\\
    & PEFNet\cite{pefnet} & 71.0 & 63.8 & -- & -- & -- & \textbf{3.6} & 63.6 & 57.2 & -- & -- & -- & \underline{1.9}\\
    & M$^2$UNet\cite{m2unet} & 76.7 & 68.4 & -- & -- & -- & \textbf{3.6} & 67.0 & 59.5 & -- & -- & -- & 2.4 \\

    \hline
    \multicolumn{2}{l}{\textbf{MEGANet(ResNet-34)}} & \underline{78.1} & \underline{70.6} & \underline{76.6} & \underline{84.5} & \textbf{89.9} & \underline{3.8} & \textbf{78.9} & \textbf{70.9} & \textbf{75.3} & \textbf{86.6} & \textbf{91.5} & \textbf{1.5} \\ 
   \multicolumn{2}{l}{\textbf{MEGANet(Res2Net-50)}} & \textbf{79.3} & \textbf{71.4} & \textbf{77.9} & \textbf{85.4} & \underline{89.5} & 4.0 & 73.9 & \underline{66.5} & \underline{70.2} & 83.6 & 85.8 & 3.7 \\
\bottomrule
\end{NiceTabular}
}
\end{table*}

\subsection{Implementation Details}
We implement MEGANet using Pytorch and an NVIDIA RTX 3090. We train our network with a batch size of 16 and a general training strategy as the ACSNet \cite{acsnet}, stochastic gradient descent (SGD) optimizer with a momentum of 0.9 and a weight decay of 1e-5. The learning rate scheduler is defined as $lr = init \ lr \times (1 {}-{} \frac{epoch}{nEpoch})^{power}$, where $init \ lr$ = 1e-3, $power$ = 0.9, $nEpoch$ = 200. We resize the input images to 352 × 352 for both the training and inference stages and then resize them back to the original size for calculating evaluation metrics. For data augmentation, we employ random flipping on both horizontal and vertical, random rotation, and a multi-scale training strategy \{0.75, 1, 1.25\}.

% As mentioned
% before, we evaluate our model in two backbones: ResNet-34 \cite{resnet} and Res2Net-50 \cite{res2net}. We adopt the pre-trained weights from the ImageNet \cite{imagenet} dataset for both backbones

\subsection{Performance Comparisons}
Corresponding to two backbone networks, ResNet-34 \cite{resnet} and Res2Net-50 \cite{res2net}, we qualitatively and quantitatively compare our MEGANet with eight SOTA methods, including U-Net \cite{unet}, U-Net++ \cite{unet++}, SFA \cite{sfa}, PraNet \cite{pranet}, SANet \cite{sanet}, MSNet \cite{msnet}, PEFNet \cite{pefnet} and M$^2$UNet \cite{m2unet}. The result of PEFNet is adapted from the M$^2$UNet, while the others are reported based on the original papers.
\begin{table*}[!h]
\centering
\caption{Performance and network efficiency comparison between our MEGANet with other existing SOTA methods on ClinicDB dataset. The highest and second highest scores are shown in \textbf{bold} and \underline{underline}, respectively. All metrics are in (\%).}\label{tab2}
\resizebox{0.8\linewidth}{!}{%
\begin{NiceTabular}{c|l|*{6}{c}|c|c}
    \toprule
    &\multirow{2}{*}{Methods} &
    \multicolumn{6}{c}{ClinicDB (seen)} &
    \multirow{2}{*}{Backbone} & 
    \multirow{2}{*}{Params(M)} \\ 
    
    \cmidrule(lr){3-8}
    & & mDice $\uparrow$ & mIoU $\uparrow$ & $F_{\beta}^{w}$ $\uparrow$ & $S_{\alpha}$ $\uparrow$ &  $E_{\phi}^{max}$ $\uparrow$ & MAE $\downarrow$\\
    \hline
    
    \multirow{8}{*}{\rotatebox[]{90}{SOTA methods}} & U-Net\cite{unet} & 82.3 & 75.5 & 81.1 & 88.9 & 95.4 & 1.9 & -- & 7.76 \\ 
    & U-Net++\cite{unet++} & 79.4 & 72.9 & 78.5 & 87.3 & 93.1 & 2.2  & -- & 9.0 \\ 
    & SFA\cite{sfa} & 70.0 & 60.7 & 64.7 & 79.3 & 88.5 & 4.2 & -- & -- \\
    & PraNet\cite{pranet} & 89.9 & 84.9 & 89.6 & 93.6 & 97.9 & 0.9  & Res2Net-50 & 32.55 \\
    & SANet\cite{sanet} & 91.6 & 85.9 & 90.9 & 93.9 & 97.6 & 1.2 & Res2Net-50 & 23.89 \\
    & MSNet\cite{msnet} & 92.1 & 87.9 & 91.4 & \underline{94.1} & 97.2 & \underline{0.8} & Res2Net-50 & 29.74 \\
    & PEFNet\cite{pefnet} & 86.6 & 81.4 & -- & -- & -- & 1.0 & EfficientNetV2-S & 27.98\\
    & M$^2$UNet\cite{m2unet} & 90.1 & 85.3 & -- & -- & -- & \underline{0.8} & MetaFormer & 28.77 \\

    \hline
    \multicolumn{2}{l}{\textbf{MEGANet(ResNet-34)}} & \underline{93.0} & \underline{88.5} & \underline{93.1} & \textbf{95.0} & \underline{98.0} & \underline{0.8} & ResNet-34 & 29.27 \\
    \multicolumn{2}{l}{\textbf{MEGANet(Res2Net-50)}} & \textbf{93.8} & \textbf{89.4} & \textbf{94.0} & \textbf{95.0} & \textbf{98.6} & \textbf{0.6} & Res2Net-50 & 44.19\\
\bottomrule
\end{NiceTabular}}
\end{table*}

\begin{table*}[h]
\caption{Impact of each component of the EGA module on model performance. The highest scores are shown in \textbf{bold}. All metrics in (\%).}
\centering
\label{ablation-ega}
\aboverulesep=0ex
\belowrulesep=0ex
\resizebox{0.8\linewidth}{!}{%
\begin{NiceTabular}{c|cccc|cc|cc|cc|cc}
\toprule
    \multirow{2}{*}{Exp.}& \multicolumn{4}{c}{EGA} & \multicolumn{2}{c}{Kvasir-SEG (seen)} & \multicolumn{2}{c}{ClinicDB (seen)}  & \multicolumn{2}{c}{ETIS (unseen)} & \multicolumn{2}{c}{CVC-300 (unseen)} \\
    \cmidrule(lr){2-5}\cmidrule(lr){6-7}\cmidrule(lr){8-9}\cmidrule(lr){10-11}\cmidrule(lr){12-13}
    & $\hat{f}^r$ & $\hat{f}^b$ & ${f}^l$ & $\mathtt{CBAM}$
    & mDice $\uparrow$ & mIoU $\uparrow$ 
    & mDice $\uparrow$ & mIoU $\uparrow$
    & mDice $\uparrow$ & mIoU $\uparrow$ 
    & mDice $\uparrow$ & mIoU $\uparrow$ \\
\hline
    \#1 & \xmark & \xmark & \xmark & \xmark & 90.0 & 84.7 & 91.7 & 86.8 & 71.6 & 63.7 & 87.1 & 80.0 \\ 
    \#2 & \xmark & \cmark & \cmark & \cmark & 90.7 & 85.3 & 92.1  & 87.5 & 76.9 & 68.7 & 86.3 & 79.4 \\
    \#3 & \cmark & \xmark & \cmark & \cmark & 90.5 & 85.3 & 92.8  & 88.3 & 78.3 & 69.9 & 88.2 & 81.5\\
    \#4 & \cmark & \cmark & \xmark & \cmark & \textbf{91.5} & \textbf{86.6} & 92.0  & 87.5 & 75.5 & 67.3 & 86.9 & 80.3	\\
    \#5 & \cmark & \cmark & \cmark & \xmark & 91.0 & 85.7 & 92.1 & 87.8 & 76.5 & 68.6 & 88.4 & 81.4\\
    \hline 
    \#6 & \cmark & \cmark & \cmark & \cmark &  91.1 & 85.9 & \textbf{93.0} & \textbf{88.5} & \textbf{78.9} & \textbf{70.9} & \textbf{88.7} & \textbf{81.8}\\
\bottomrule
\end{NiceTabular}}
\end{table*}

\begin{figure*}[h]
\setkeys{Gin}{width=\linewidth,height=0.75\linewidth}
\renewcommand{\arraystretch}{0.3}
\setlength{\tabcolsep}{1pt}
\aboverulesep=0ex
\belowrulesep=0ex
\begin{tabularx}{\textwidth}{aX|XXXX|XX|X}  

    & \centering Image & \centering SFA & \centering PraNet & \centering SANet & \centering MSNet & \multicolumn{2}{X|}{\enspace\enspace\enspace\enspace\enspace\enspace\enspace\enspace\boldblue{MEGANet}}& \centering \boldred{GT} \endline
    
    & & & & & & \centering \textbf{ResNet-34} & \centering \textbf{Res2Net-50} & \\
    
     \raisebox{3.0\height}{a1}& \includegraphics{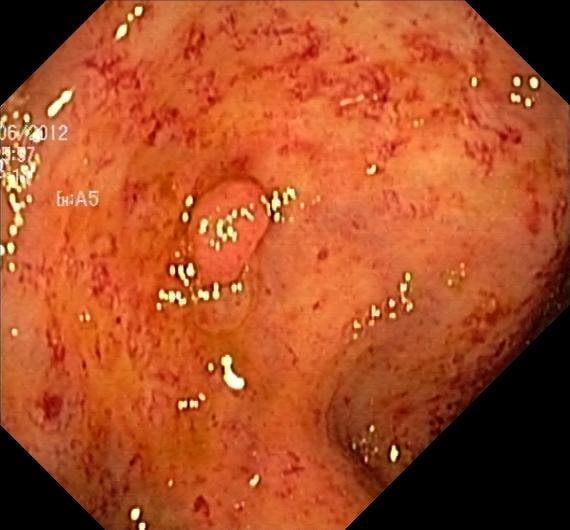} & \includegraphics{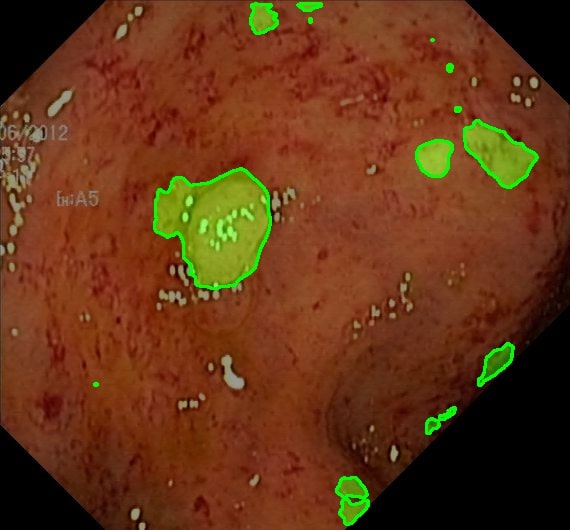} & \includegraphics{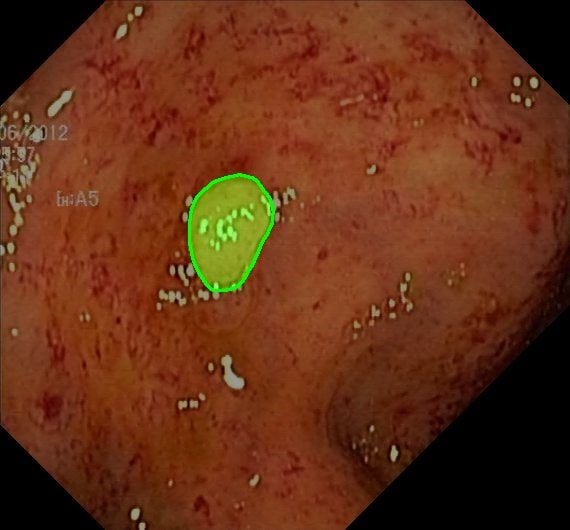} & \includegraphics{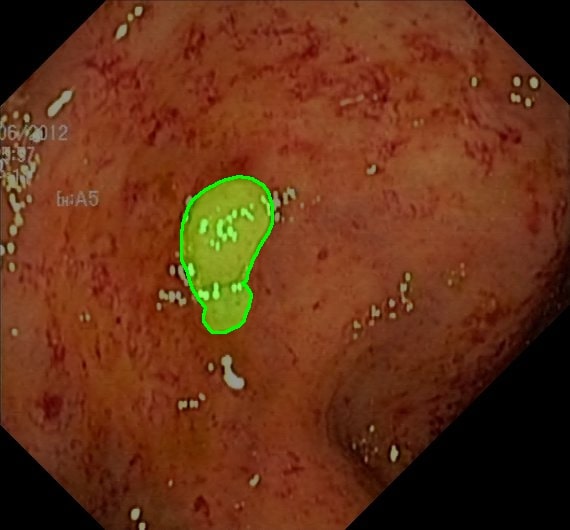} & \includegraphics{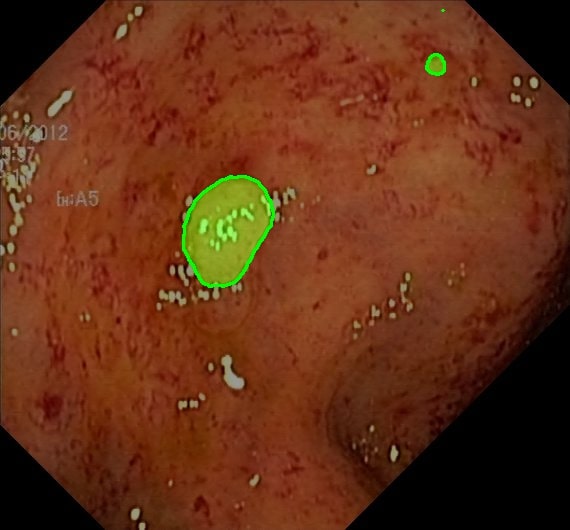} & 
    \includegraphics{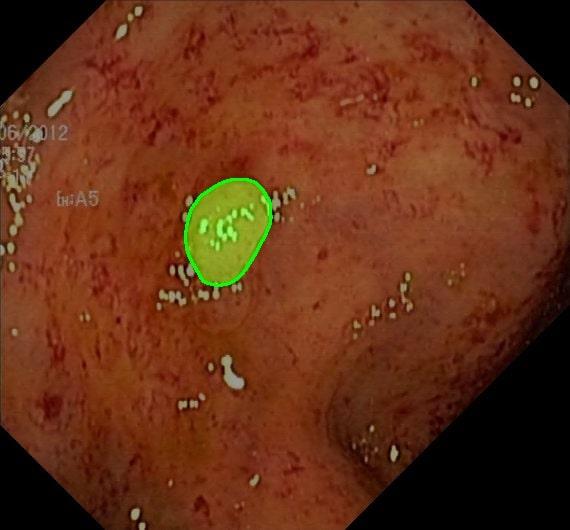} & 
    \includegraphics{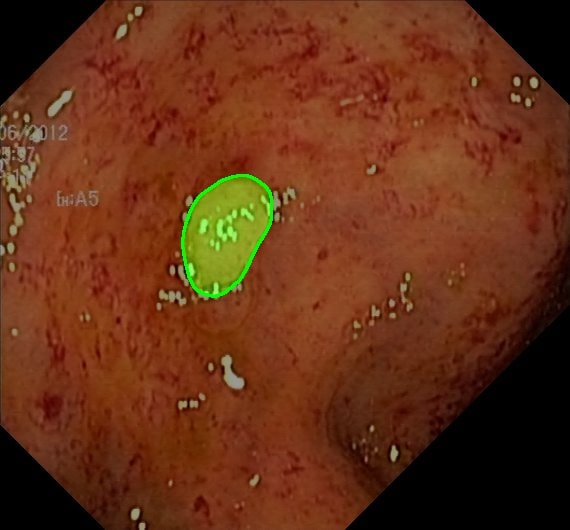} & \includegraphics{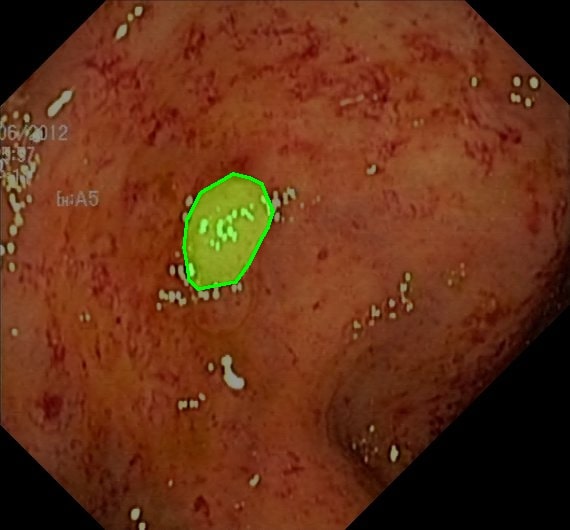} \\

    \raisebox{3.0\height}{a2}& 
    \includegraphics{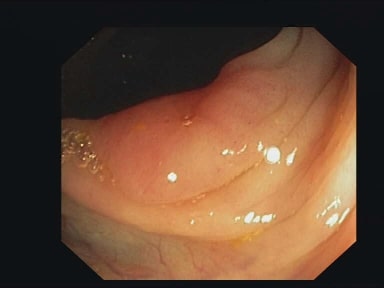} &
    \includegraphics{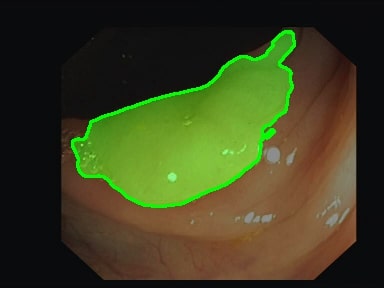} &
    \includegraphics{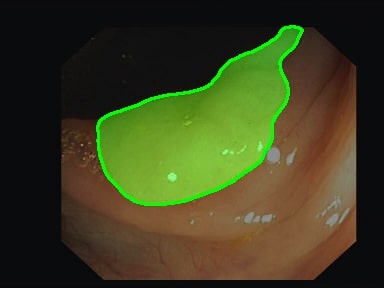} &
    \includegraphics{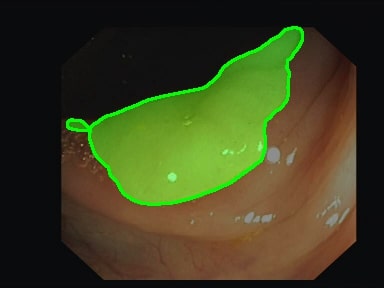} & 
    \includegraphics{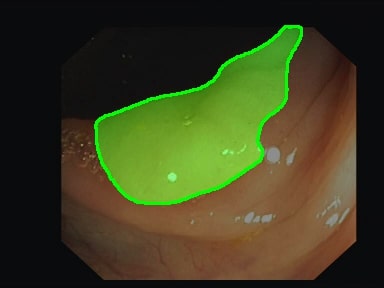} & 
    \includegraphics{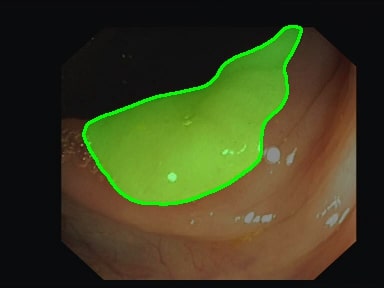} &
    \includegraphics{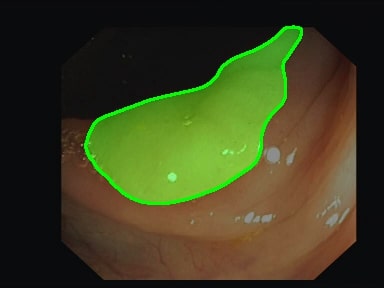} & 
    \includegraphics{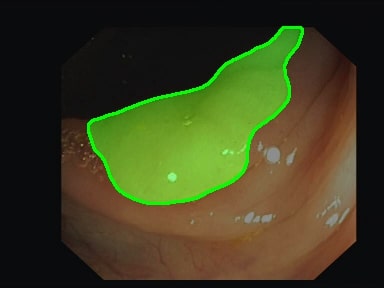} \\

    \raisebox{3.0\height}{b1}& 
    \includegraphics{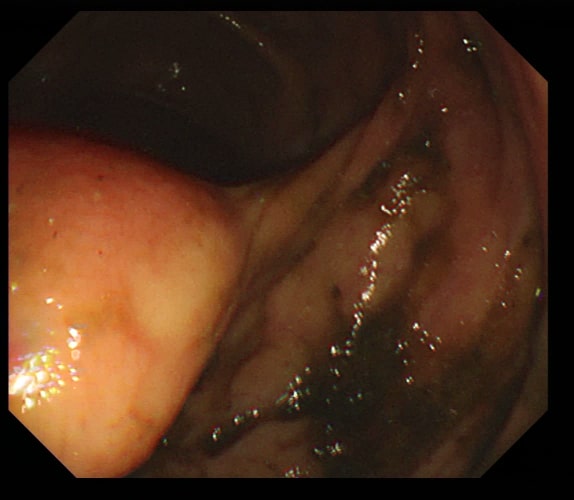} &
    \includegraphics{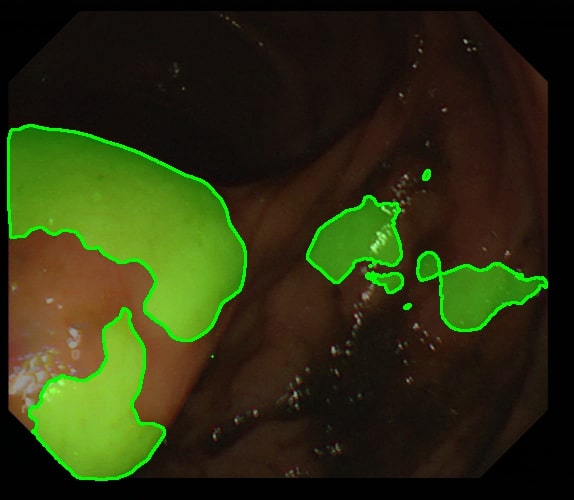} &
    \includegraphics{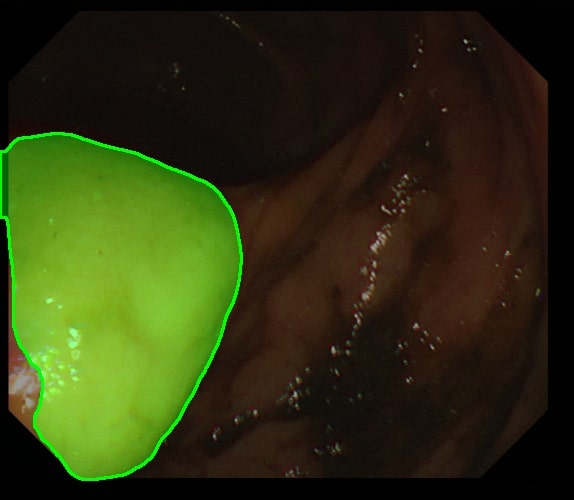} &
    \includegraphics{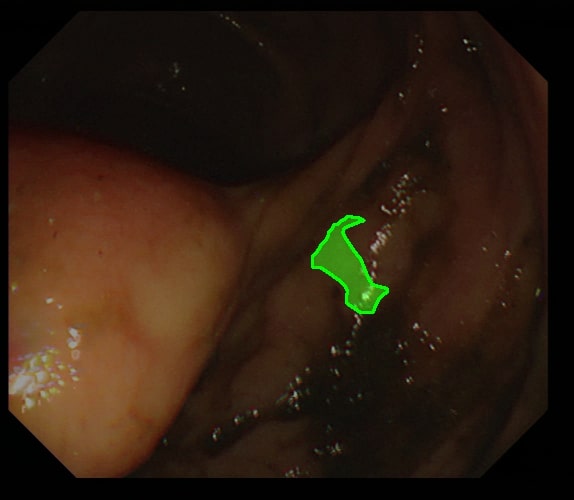} &
    \includegraphics{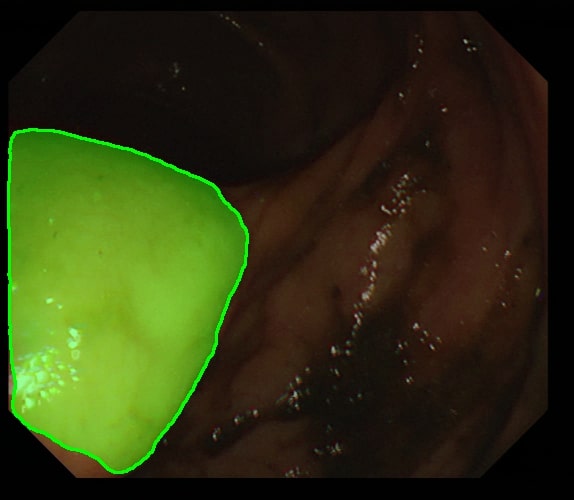} & 
    \includegraphics{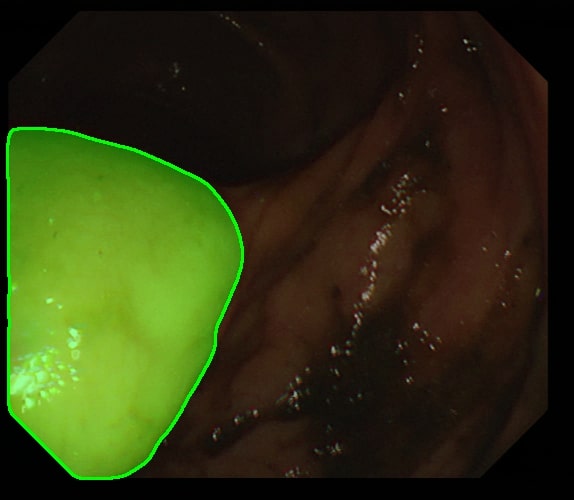} & 
    \includegraphics{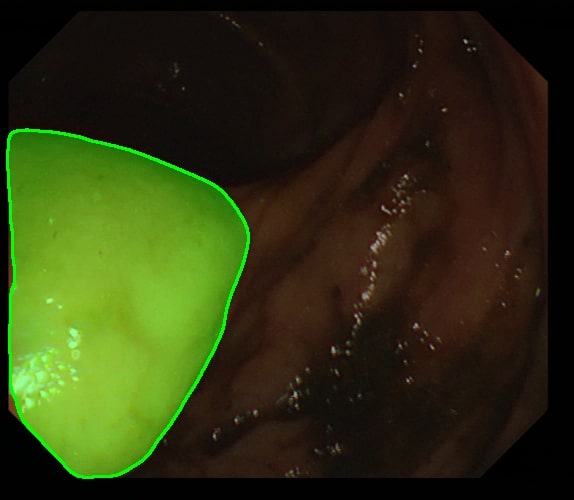} & 
    \includegraphics{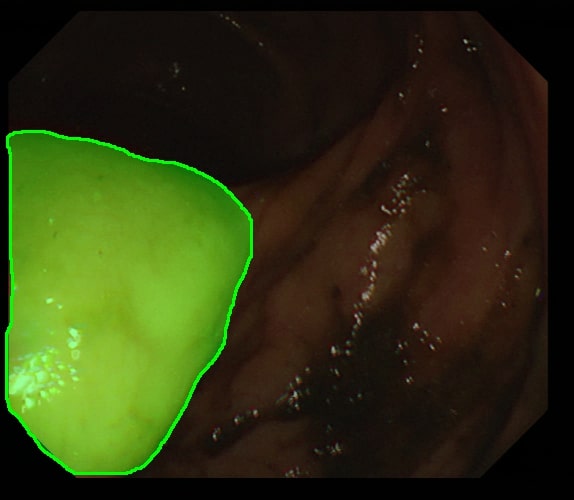} \\

    \raisebox{3.0\height}{b2}& 
    \includegraphics{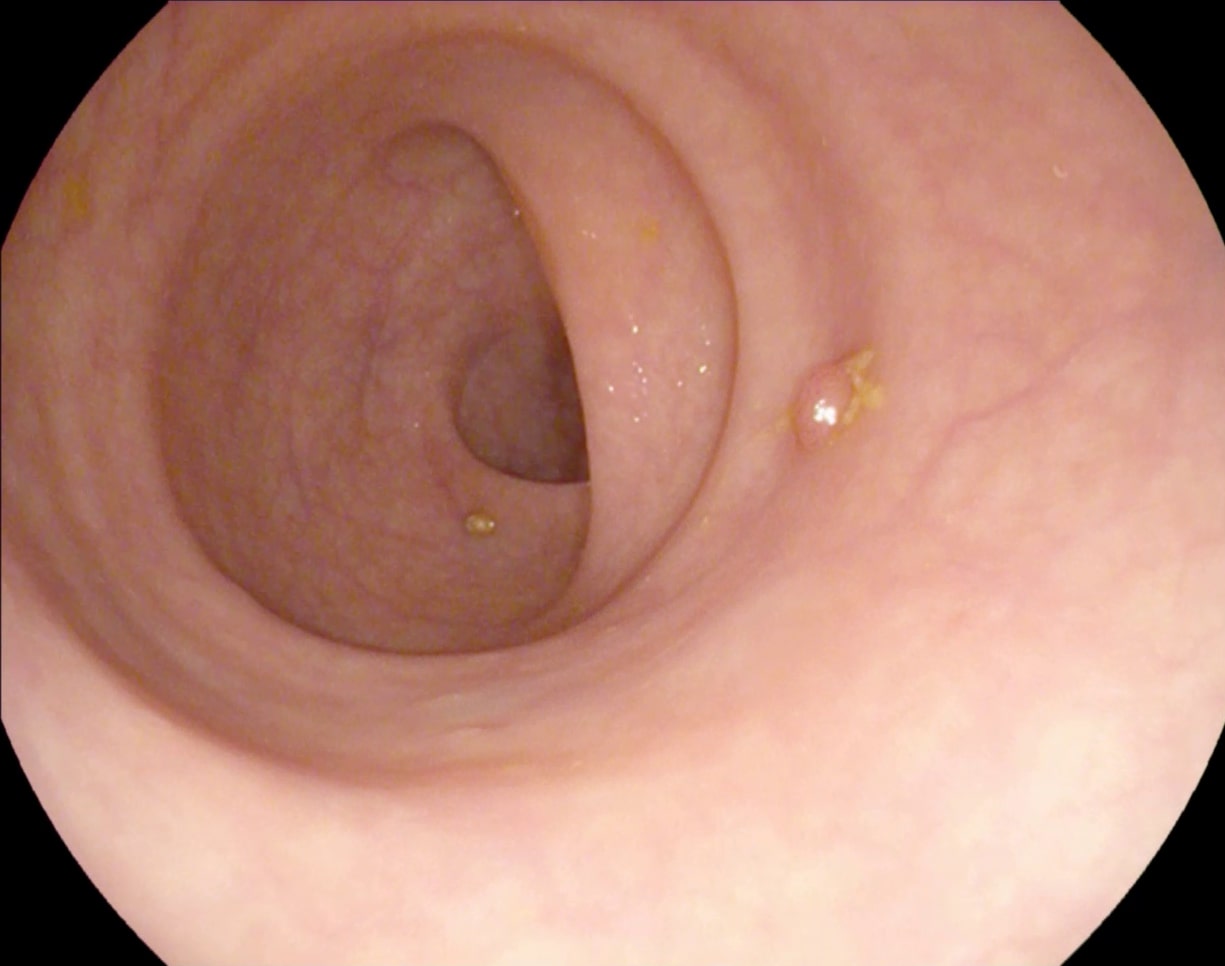} & 
    \includegraphics{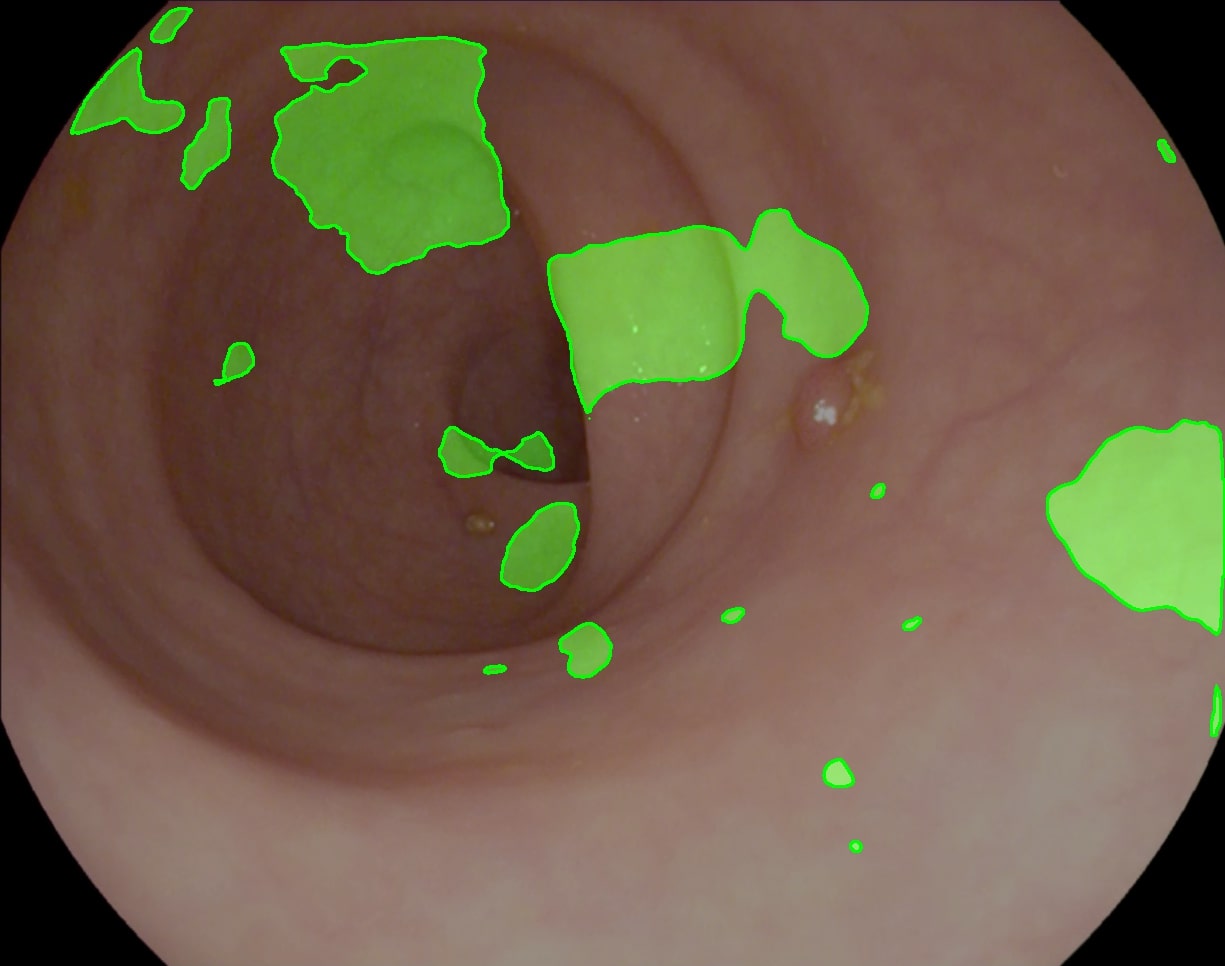} & 
    \includegraphics{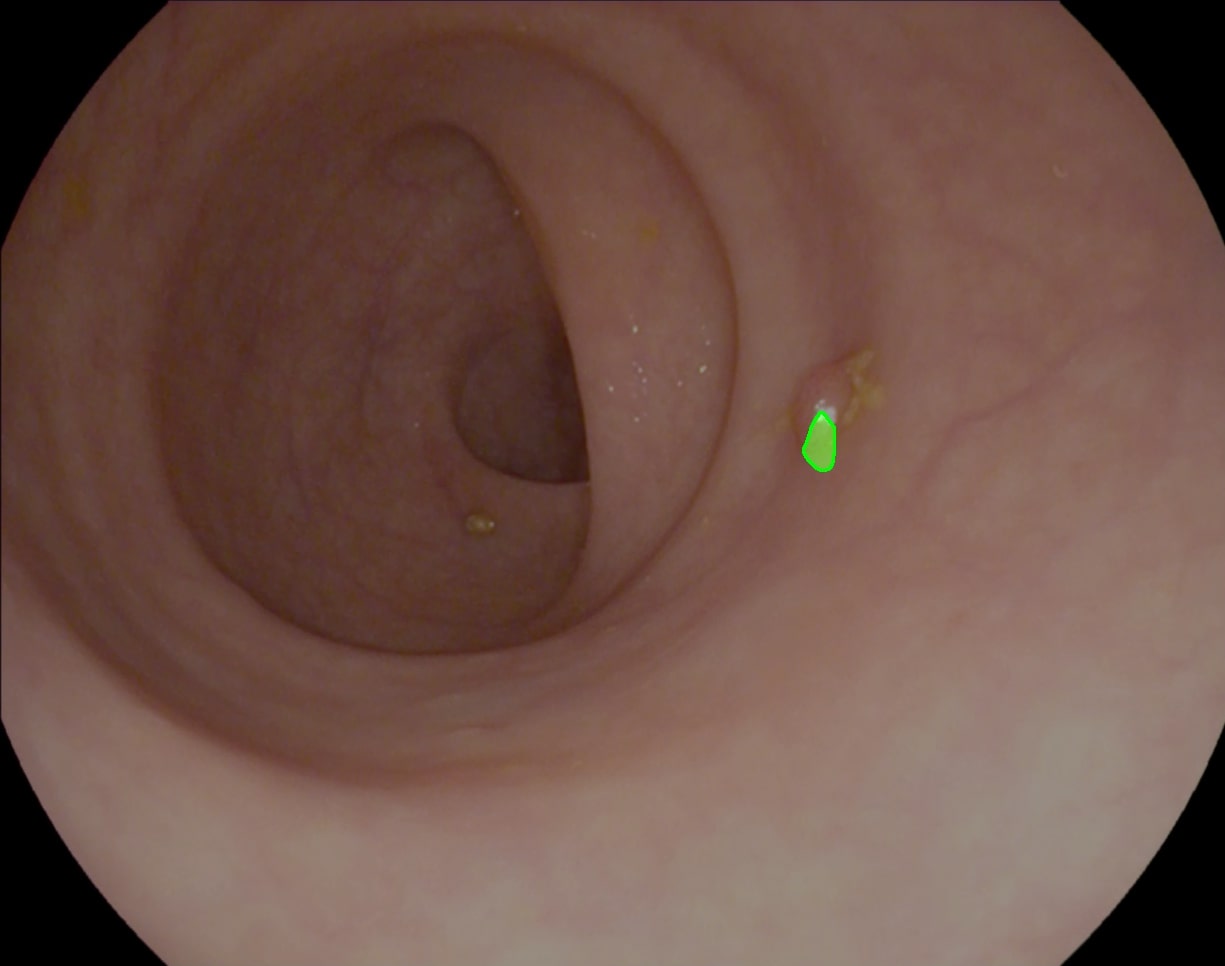} &
    \includegraphics{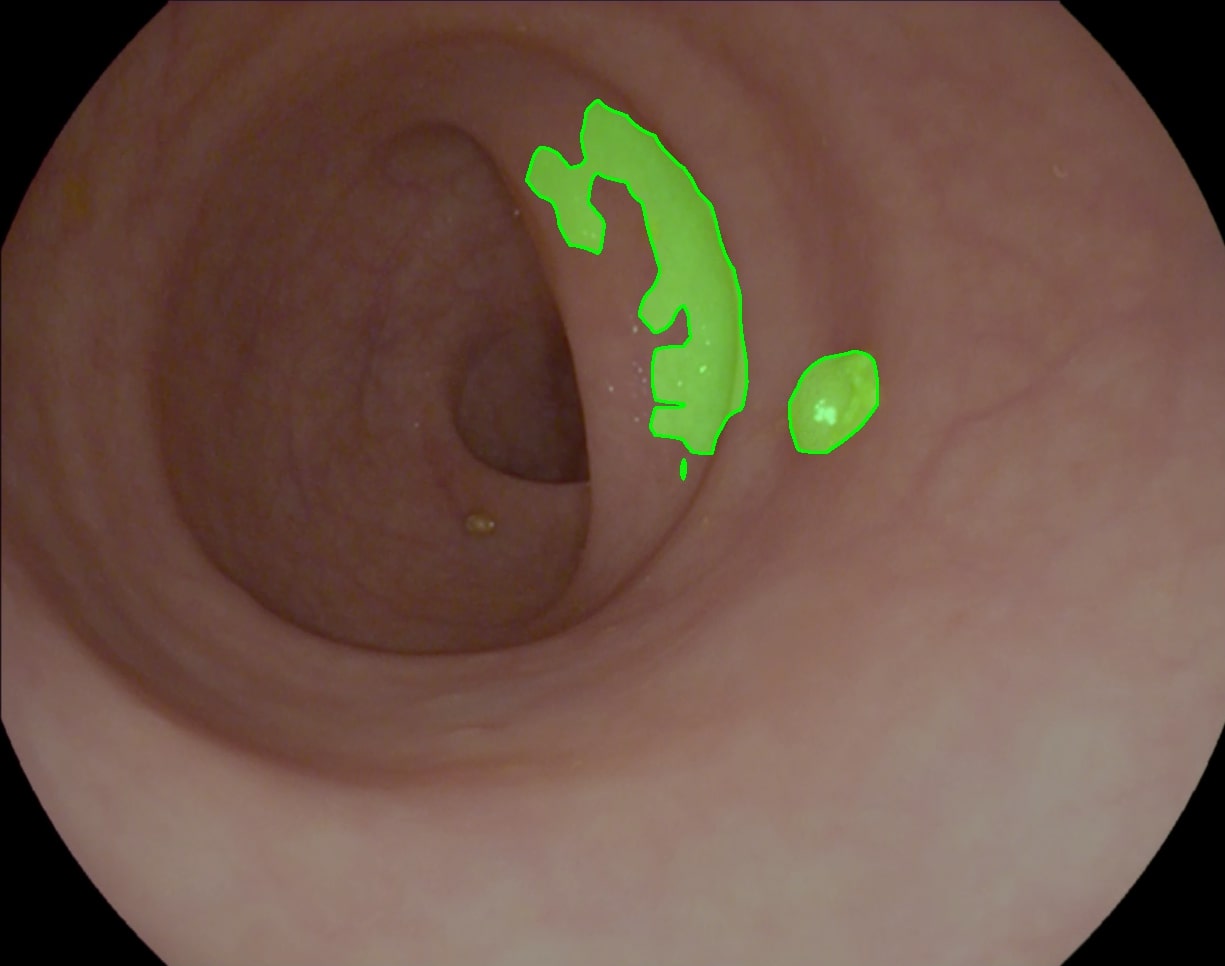} & 
    \includegraphics{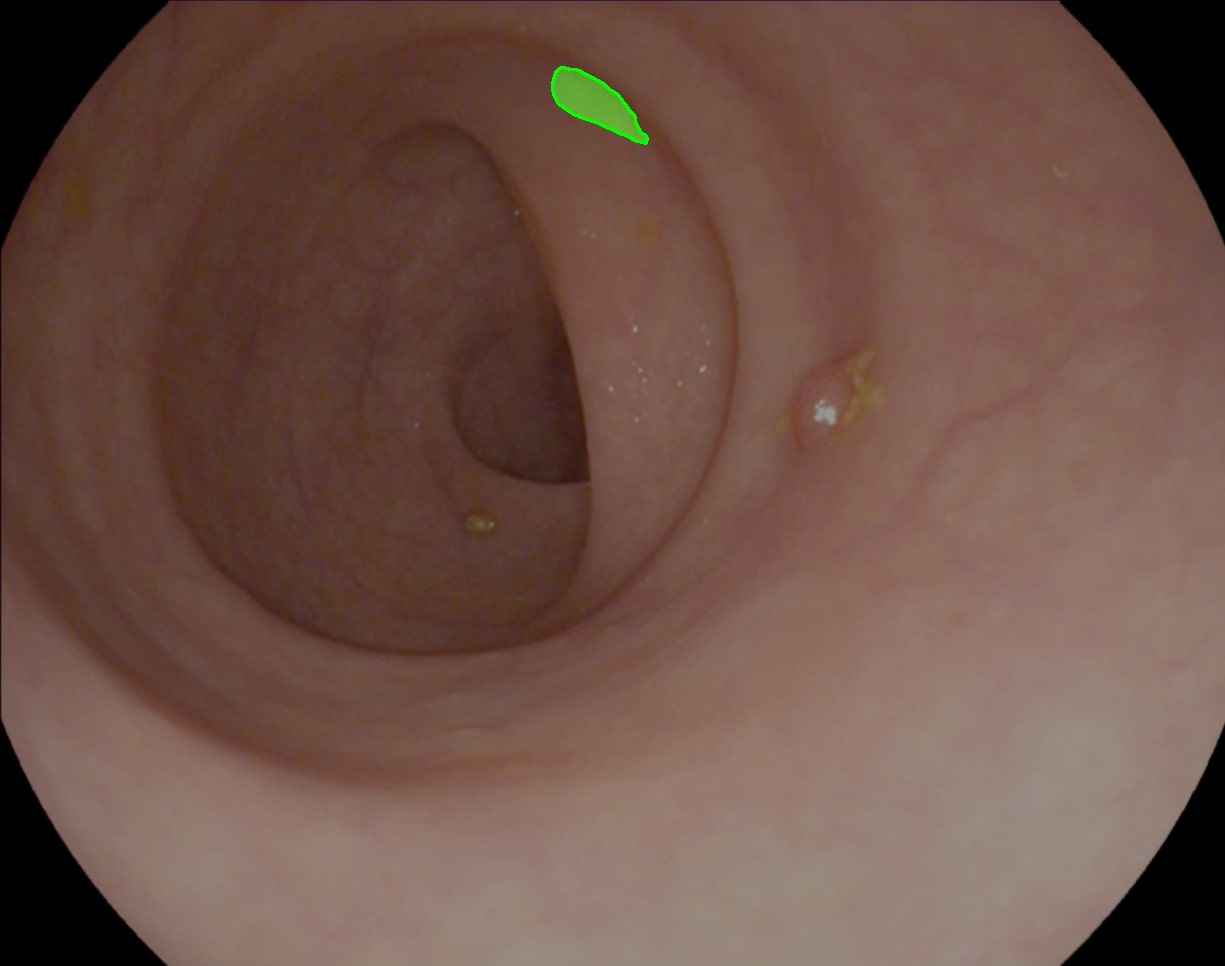} & 
    \includegraphics{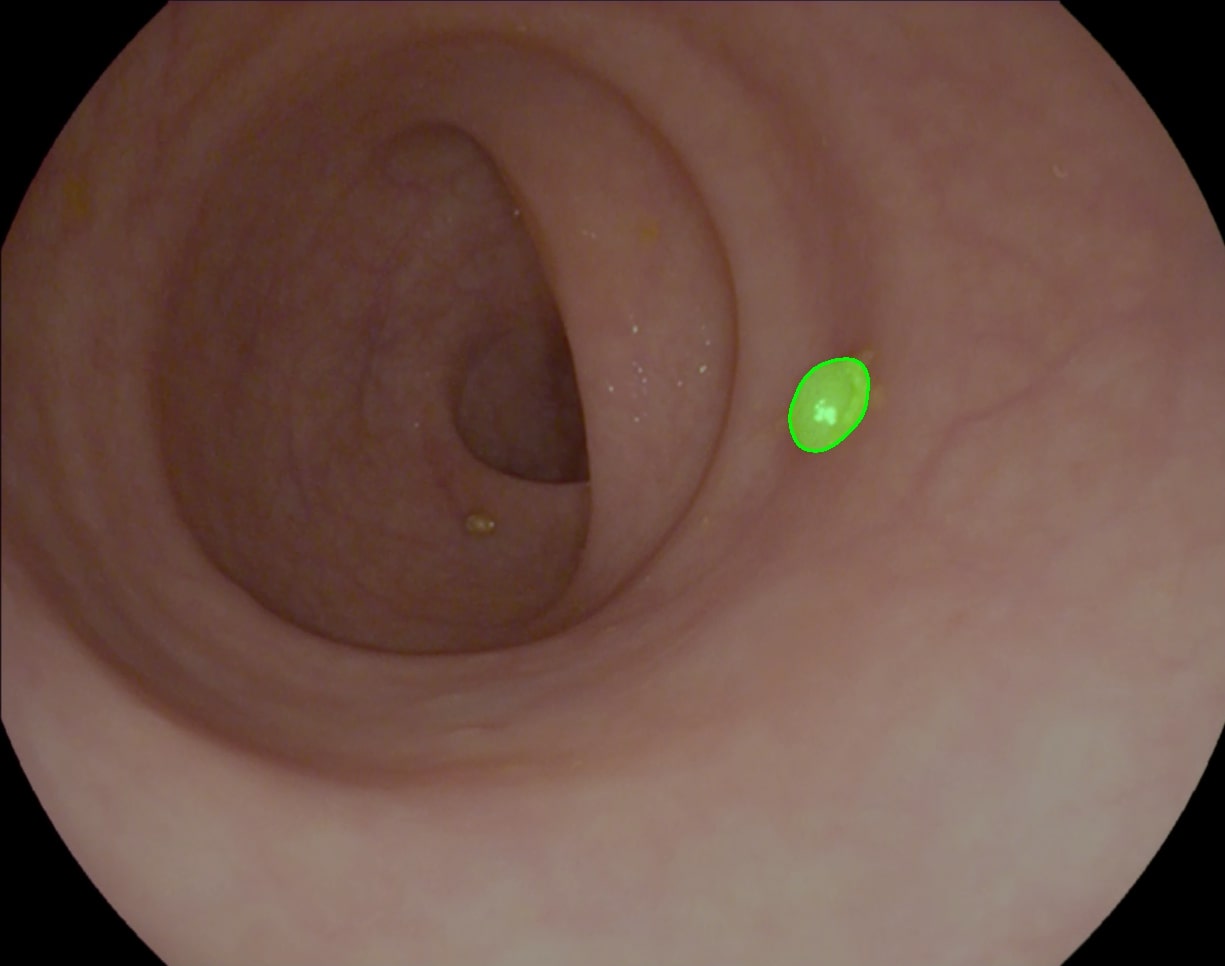} & 
    \includegraphics{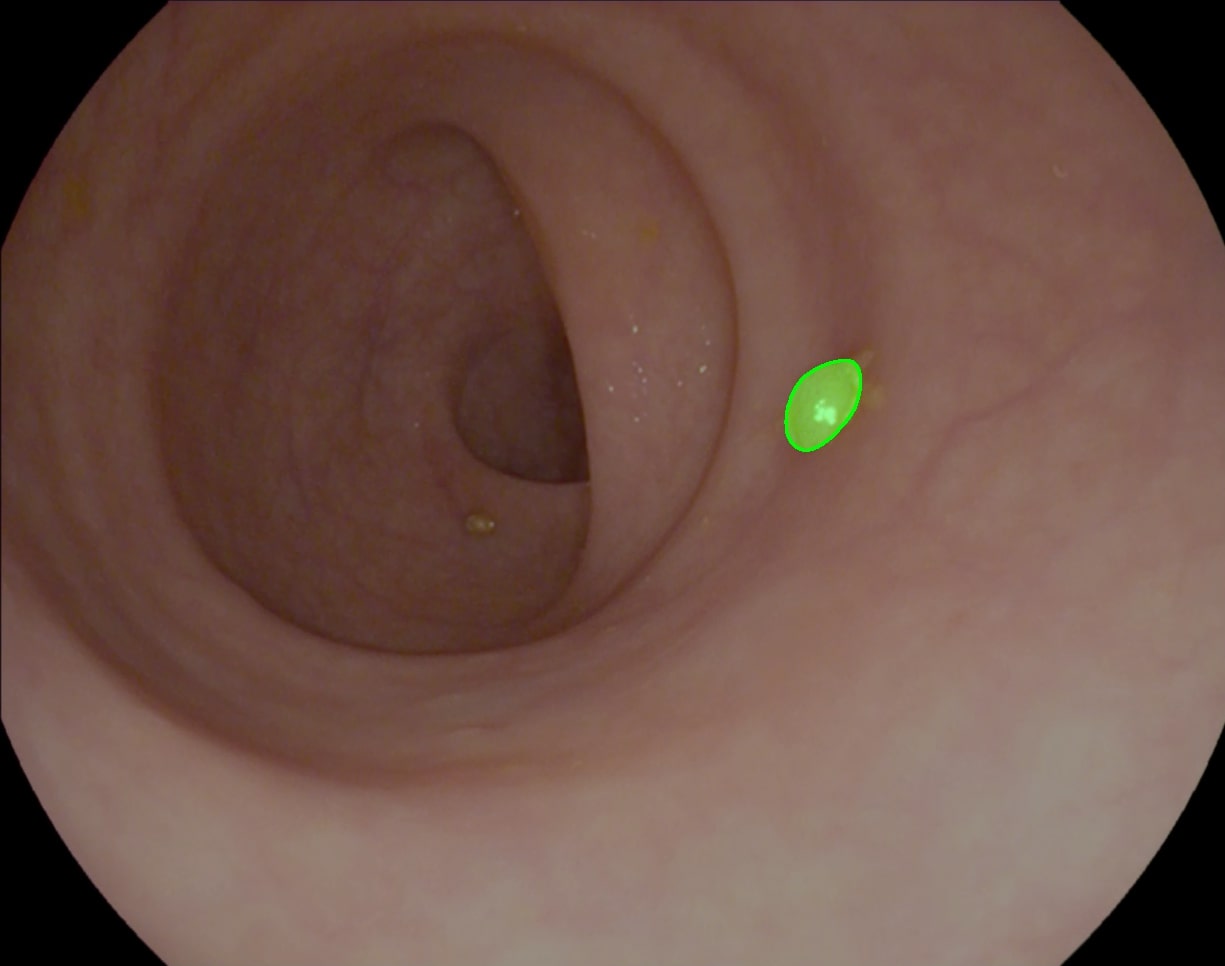} & 
    \includegraphics{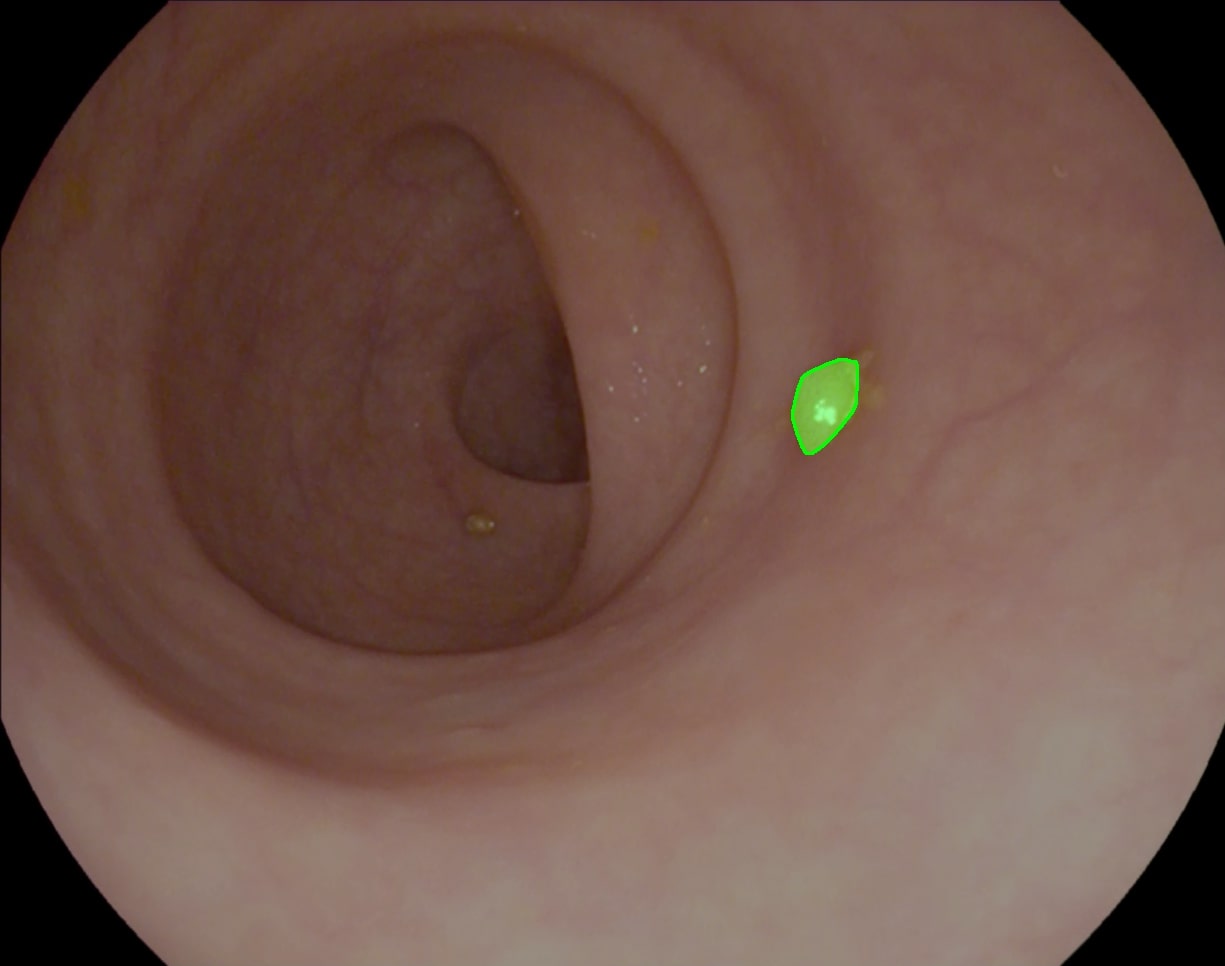} \\

    \raisebox{3.0\height}{c1}& 
    \includegraphics{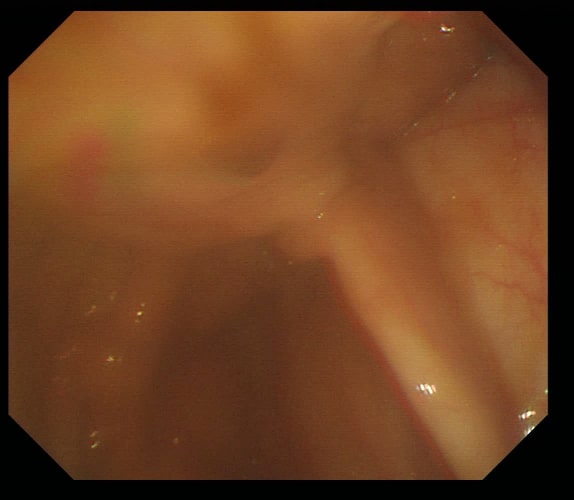}&
    \includegraphics{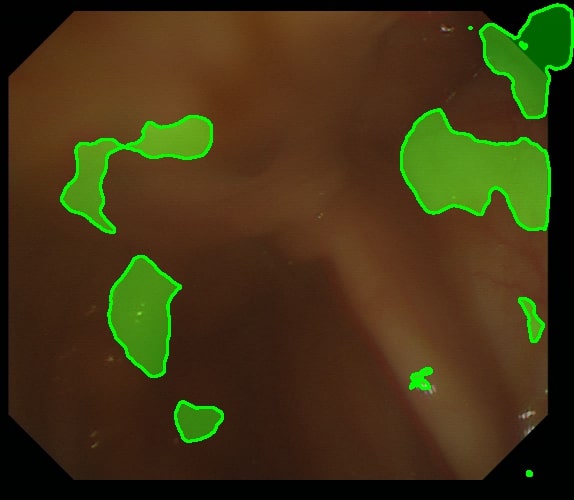}&
    \includegraphics{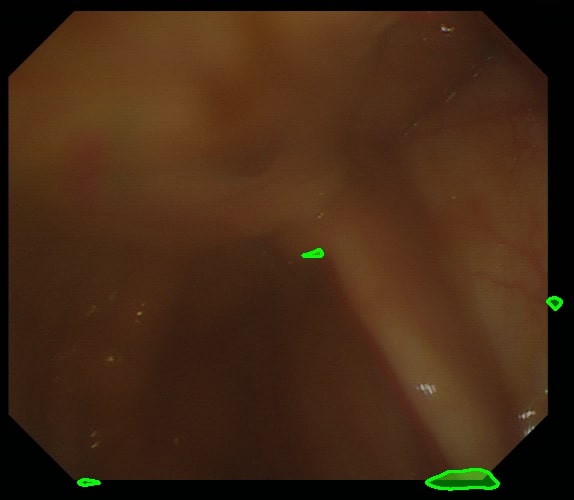}&
    \includegraphics{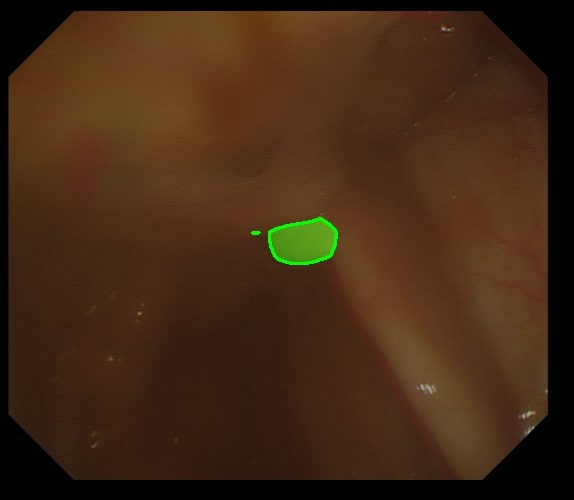}&
    \includegraphics{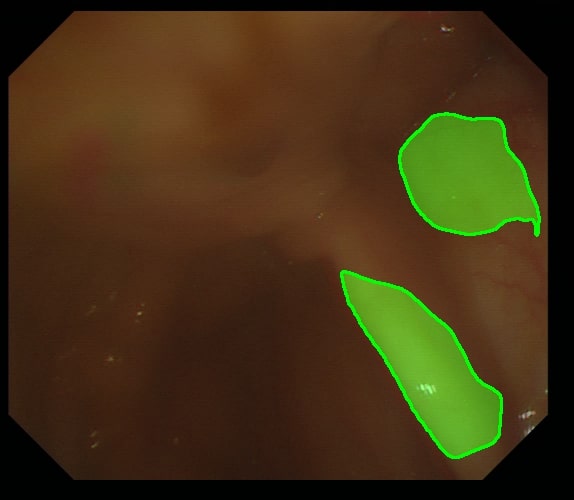}&
    \includegraphics{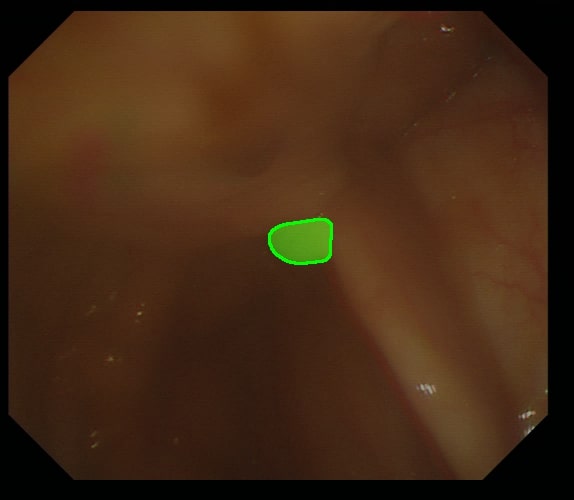}&
    \includegraphics{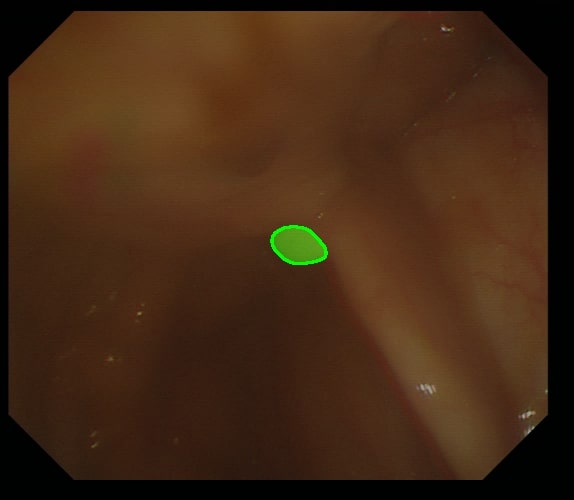}&
    \includegraphics{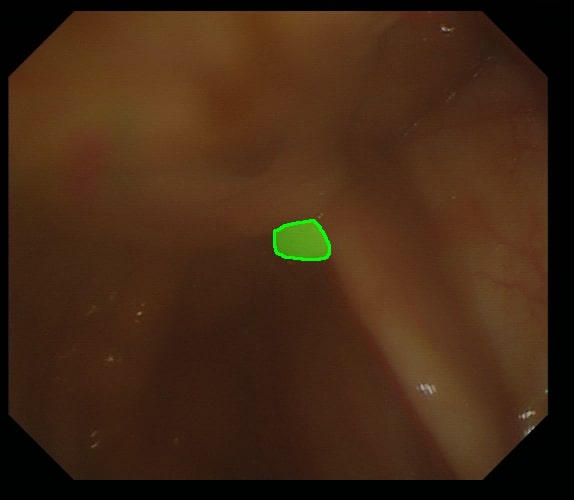}\\

    \raisebox{3.0\height}{c2}&
    \includegraphics{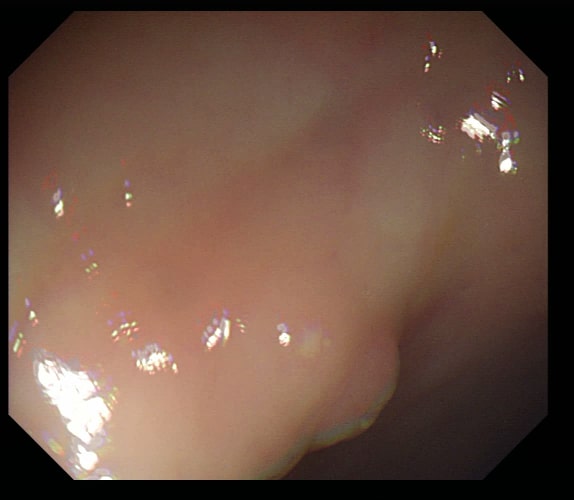}&
    \includegraphics{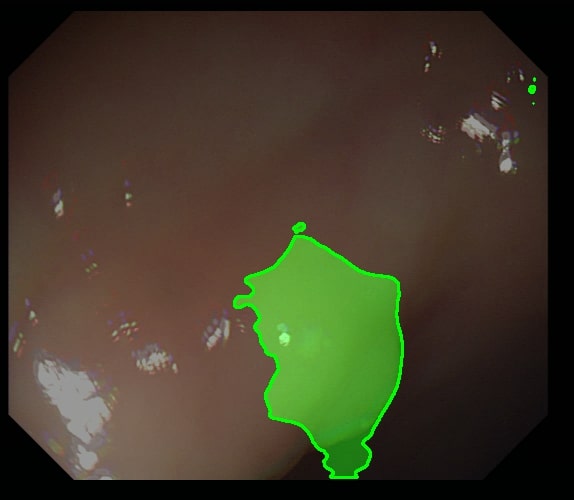}&
    \includegraphics{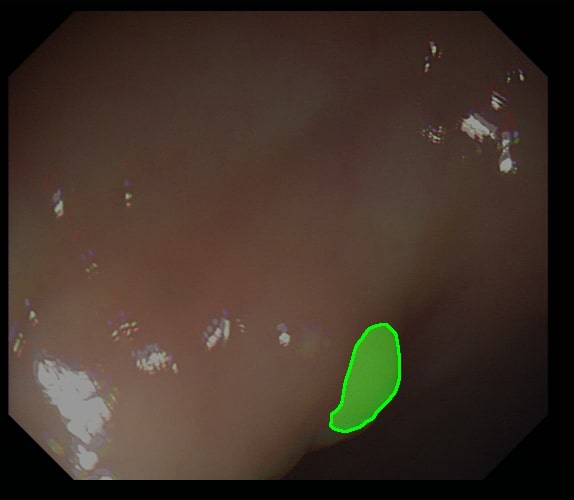}&
    \includegraphics{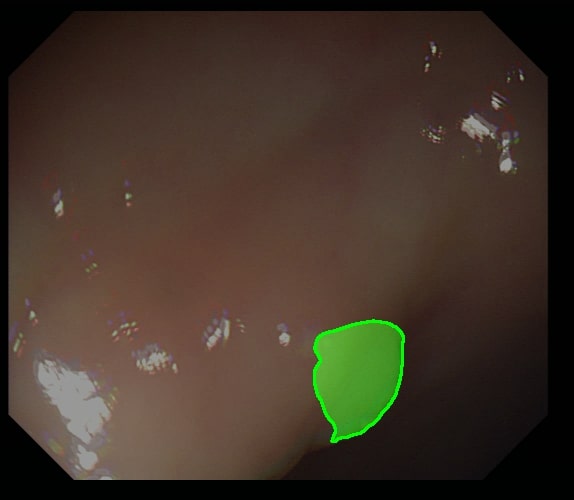}&
    \includegraphics{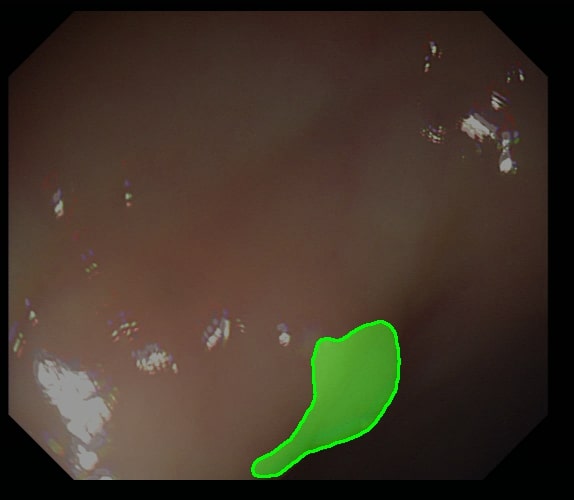}&
    \includegraphics{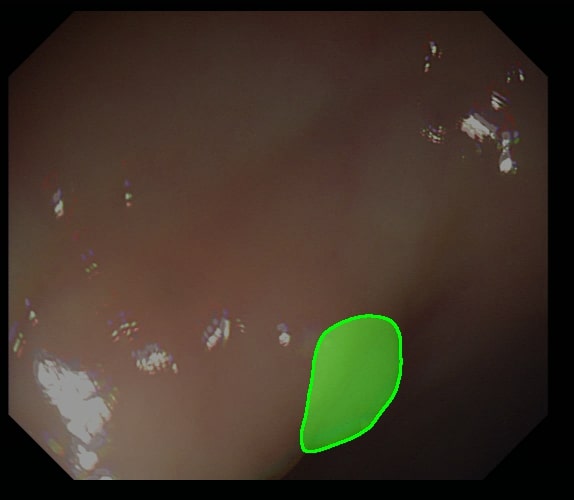}&
    \includegraphics{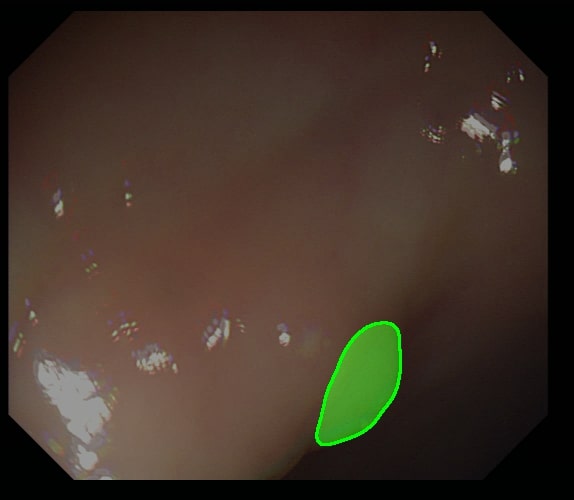}&
    \includegraphics{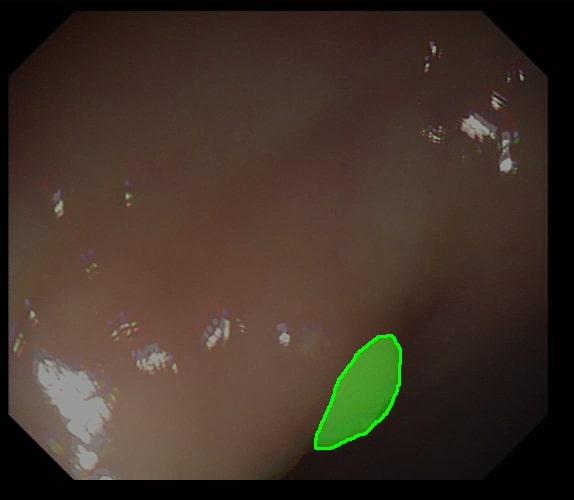}\\

\end{tabularx}
\vspace{-0.01in}
\caption{Qualitative comparison between our MEGANet and existing SOTA methods. The a1 image comes from the Kvasir dataset, while the a2 is from ClinicDB. The b2 image is derived from the ETIS dataset, and the rest are from the ColonDB dataset.} 
\vspace{-0.01in}
\label{qualitative}
\end{figure*}
\noindent
\textbf{Quantitative Evaluation.}
Table \ref{tab1} presents a quantitative performance comparison between our MEGANet and other SOTA methods on the Kvasir-SEG, CVC-300 (a subset of EndoScene), ColonDB, and ETIS datasets. We provide the corresponding performance results for both MEGANet backbones, namely ResNet-34 and Res2Net-50. Like other SOTA methods, our models were trained on the Kvasir-SEG and CVC-ClinicDB training sets. The performance metrics reported on the Kvasir-SEG dataset are classified as seen, while those reported on CVC-300, ColonDB, and ETIS datasets are considered unseen. From the insights presented in Table \ref{tab1}, it is evident that MEGANet (Res2Net-50) excels in numerous metrics, demonstrating superior performance. Especially, our MEGANet with the backbone ResNet-34 leads the other methods by remarkable gaps on most metrics when tested on the ETIS dataset. On investigating this, we observe that the polyps in the ETIS dataset are smaller than the others. Thanks to the bounded capacity of the ResNet-34 \cite{resnet} backbone, MEGANet with this backbone is prone to avoid overfitting, achieving preferable results on this dataset.

In addition to the performance evaluation, we also factor in network efficiency for a comprehensive comparison, as outlined in Table \ref{tab2}. This evaluation is conducted specifically on the ClinicDB dataset. Consider the example of M$^2$UNet \cite{m2unet}: our MEGANet (ResNet-34) possesses a comparable number of network parameters, yet it outperforms M$^2$UNet with a significant 2.9\% improvement in mDice and a 3.2\% enhancement in mIOU. Furthermore, when contrasted with all existing SOTA methods, our MEGANet (Res2Net-50) obtains the best performance across all metrics, even with a relatively small number of network parameters.

\noindent
\textbf{Qualitative Evaluation.} Visual comparisons of each polyp segmentation challenge are illustrated in Figure \ref{qualitative}. In particular, (a1 and a2) highlight the complex background issue, (b1 and b2) exemplify the variability in polyp sizes and configurations, while (c1 and c2) demonstrate the challenge of dealing with indistinct boundaries. Notably, our approach reflects the capability to address various sizes and shapes within each challenge. Particularly, the results in the case of (b2 and c1) underscore the exceptional performance of our method in maintaining an impressively low false positive rate, accurately refraining from misclassifying healthy regions as tumors.

\begin{table*}[h]
\caption{Ablation study to assess the methodology for computing high-frequency features $f^l_i$ with two formulations: our high-frequency feature (Equation \ref{eq:fl}) and Laplacian pyramid features (Equation \ref{eq:fl2}). The highest scores are shown in \textbf{bold}. All metrics in (\%).}
\label{ablation-pyramid}
\centering
\aboverulesep=0ex
\belowrulesep=0ex
\begin{tabular}{c|cc|cc|cc|cc}
\toprule
    \multirow{2}{*}{$f^l_i$} & \multicolumn{2}{c|}{Kvasir-SEG (seen)} & \multicolumn{2}{c|}{ClinicDB (seen)}  & \multicolumn{2}{c|}{ETIS (unseen)} & \multicolumn{2}{c}{CVC-300 (unseen)} \\
    \cmidrule(lr){2-3}\cmidrule(lr){4-5}\cmidrule(lr){6-7}\cmidrule(lr){8-9}
    & mDice $\uparrow$ & mIoU $\uparrow$ 
    & mDice $\uparrow$ & mIoU $\uparrow$
    & mDice $\uparrow$ & mIoU $\uparrow$ 
    & mDice $\uparrow$ & mIoU $\uparrow$ \\
\hline
     Equation \ref{eq:fl} & \textbf{91.1} & \textbf{85.9} & \textbf{93.0} & \textbf{88.5} & \textbf{78.9} & \textbf{70.9} & \textbf{88.7} & \textbf{81.8}\\
    Equation \ref{eq:fl2} & 89.9 & 85.8 & 92.6 & 88.1 & 78.0 & 70.2 & 88.0 & 81.4\\ 
\bottomrule
\end{tabular}
\end{table*}

\subsection{Ablation Study}
As previously mentioned, the EGA module is composed of three key components: the encoded visual feature $\hat{f}^e$, the decoded predicted feature $\hat{f}^d$, which is further decomposed into reverse attention $\hat{f}^r$, boundary attention $\hat{f}^b$, and the high-frequency feature $f^l$. Additionally, the EGA incorporates the $\mathtt{CBAM}$ module. To assess the effectiveness of each input component ($\hat{f}^r$, $\hat{f}^b$, $f^l$) and the $\mathtt{CBAM}$ module within the EGA, we systematically remove each of these components as well as $\mathtt{CBAM}$, conducting an ablation study on both seen datasets (Kvasir and ClinicDB) and unseen datasets (ETIS and CVC-300). The results are presented in Table \ref{ablation-ega}.

Based on the empirical findings, each component within the EGA framework distinctly contributes to enhancing predictive performance. Comparing \#5 to \#6, we can observe the impact of the $\mathtt{CBAM}$ component. Focusing on \#1 and \#5, we can discern the effect of the combined use of $\hat{f}^r$, $\hat{f}^b$, and $f^l$. The comparison between \#2 and \#4 highlights the influence of the combined utilization of $\hat{f}^r$ and $\hat{f}^b$, which is also indicative of the influence of $\hat{f}^d$. Experiments \#2 and \#3 specifically isolate the effects of $\hat{f}^r$ and $\hat{f}^b$, respectively. Notably, the comparison between \#4 and \#6 underscores the pivotal role played by the high-frequency feature ${f}^l$. It's essential to note that the Kvasir dataset's mucous membrane (background) presents a highly intricate composition, causing the high-frequency feature ${f}^l$ of the input image to contain noise. Consequently, the version without ${f}^l$ attains the highest score within the Kvasir dataset.

We also conducted an ablation experiment to assess the methodology for computing high-frequency features $f^l_i$ computation, as defined in Equation \ref{eq:fl} by comparing it with high-frequency features $f^l_i$ obtained from Laplacian pyramid (Equation \ref{eq:1}, \ref{eq:2}). In other words, we compare the performance of MEGANet when using Equation \ref{eq:fl}  and the following equation
\begin{equation}
    f^l_i = L_i(I) = I_i - u(I_{i+1}) =  I_i - u[d(g(I_i)]
\label{eq:fl2}
\end{equation}
Table \ref{ablation-pyramid} presents the results of two scenarios: using our proposed Equation \ref{eq:fl} and obtaining the features from the Laplacian pyramid as described in Equation \ref{eq:fl2}.

% , which is shown in Table \ref{ablation-pyramid}. Particularly, we compare the performance when employing all levels of the Laplacian pyramid with the performance when just bilinearly resizing the first feature $f^l$. We claim that resizing gives better results because higher levels of the Laplacian pyramid discard relevant information while bilinear downsampling retains it.

% \noindent
% \textbf{\large Conclusion}

\section{Conclusion}
This paper introduces a novel approach called Multi-Scale Edge-Guided Attention Network (MEGANet) for polyp segmentation. The key innovation is the integration of the Edge-Guided Attention (EGA) module, designed to retain crucial high-frequency details (such as edges) to enhance the detection of weak boundary polyp objects. Our EGA module at the $i^{th}$ level amalgamates information from the $i^{th}$ layer encoder, the $i^{th}$ layer's high-frequency component, and the $(i+1)^{th}$ layer decoder. To maintain the integrity of high-frequency information, we propose deriving the high-frequency component from the base level rather than applying Gaussian filtering at each layer. Experimental results underscore the effectiveness of our MEGANet in polyp segmentation. The assessment is based on a range of metrics, including localization measures (mDice, mIoU), accuracy ($F^w_\beta$), and structural assessments ($S_{\alpha}$, $E_{\phi}^{max}$, MAE), all of which demonstrate the advantages of our proposed MEGANet. 

% To preserve the structure boundary of the polyp and enhance the distinguishable ability between foreground and background, we perform the Laplacian operator from the original image as well as the predicted map and combine it with the reverse attention at several scales by the simple attention mechanism. Based on the experiments, our method can solve various challenges, improve the performance and the generalization of unseen datasets for the polyp segmentation task. Since the EGA module can be incorporated into any UNet-based network, it is also interesting to adapt EGA to other networks to demonstrate the ability of our module.

% \section*{Acknowledgement}
\noindent
\textbf{Acknowledgement.}
Nhat-Tan Bui and Ngan Le are sponsored by the National Science Foundation (NSF) under Award No OIA-1946391 RII Track-1, NSF 1920920 RII Track 2 FEC, NSF 2223793 EFRI BRAID, NSF 2119691 AI SUSTEIN, NSF 2236302, NIH 1R01CA277739-01.
Dinh-Hieu Hoang and Quang-Thuc Nguyen are funded by Vingroup Joint Stock Company and supported by the Domestic Master/ PhD Scholarship Programme of Vingroup Innovation Foundation (VINIF), Vingroup Big Data Institute (VINBIGDATA), code VINIF.2022.ThS.JVN.04 and VINIF.2022.ThS.JVN.10, respectively. 
Minh-Triet Tran is sponsored by Vietnam National University Ho Chi Minh City (VNU-HCM) under grant number DS2020-42-01.

% \section*{Acknowledgment}
% This material is based upon work supported by the National Science Foundation (NSF) under Award No OIA-1946391 RII Track-1, NSF 1920920 RII Track 2 FEC, NSF 2223793 EFRI BRAID, NSF 2119691 AI SUSTEIN, NSF 2236302.

%%%%%%%%% REFERENCES
{
\bibliographystyle{ieee_fullname}
\bibliography{egbib}
}
% \clearpage

% \begin{figure*}[ht]
% \centering
%     \includegraphics[width=1\textwidth, page=1]{Supplementary_Polyp.pdf}
% \end{figure*}

% \begin{figure*}[ht]
% \centering
%     \includegraphics[width=1\linewidth, page=2]{Supplementary_Polyp.pdf}
% \end{figure*}

% \begin{figure*}[ht]
% \centering
%     \includegraphics[width=1\linewidth, page=3]{Supplementary_Polyp.pdf}
% \end{figure*}
\end{document}

\title{MEGANet: Multi-Scale Edge-Guided Attention Network\\ for Weak Boundary Polyp Segmentation \\
(Supplementary Material)}

\author{Nhat-Tan Bui$^{1}$, Dinh-Hieu Hoang$^{2, 3}$, Quang-Thuc Nguyen$^{2, 3}$, Minh-Triet Tran$^{2, 3}$, Ngan Le$^{1}$\\
$^1$University of Arkansas, Fayetteville, Arkansas, USA \\
$^2$University of Science, and John von Neumann Institute, VNU-HCM  \\
$^3$Vietnam National University, Ho Chi Minh City, Vietnam \\
}

\maketitle

\section{Visualization of the level-1 Laplacian pyramid}

\begin{figure*}[t]
\setkeys{Gin}{width=\linewidth,height=\linewidth}
\renewcommand{\arraystretch}{0.3}
\setlength{\tabcolsep}{1pt}
\begin{center}
\begin{tabularx}{\textwidth}{sXXXXX}  
    
    \raisebox{5.0\height}{Image}& 
    \includegraphics{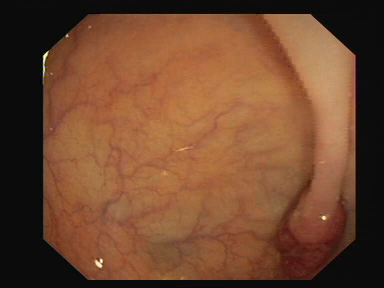} & 
    \includegraphics{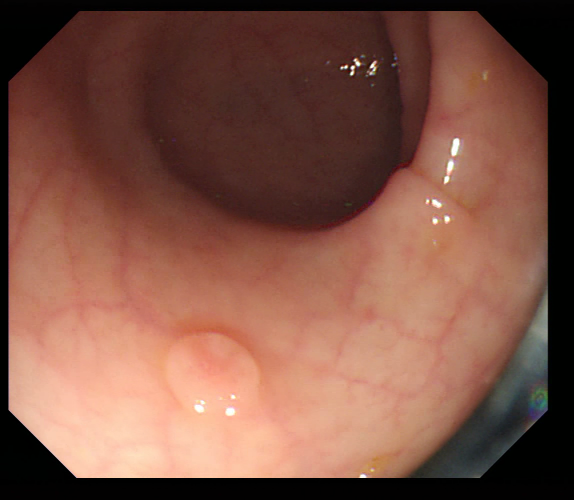} &
    \includegraphics{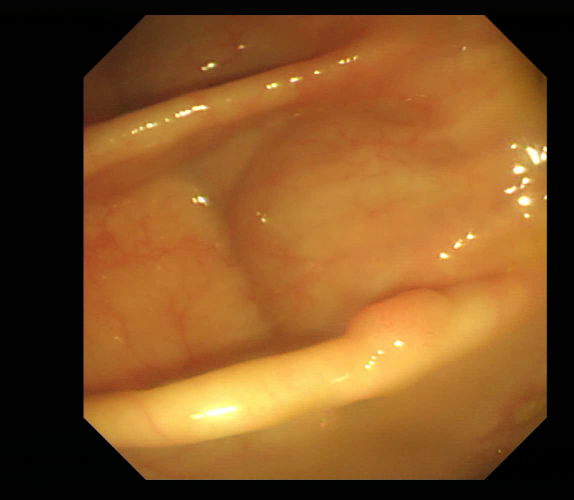} &
    \includegraphics{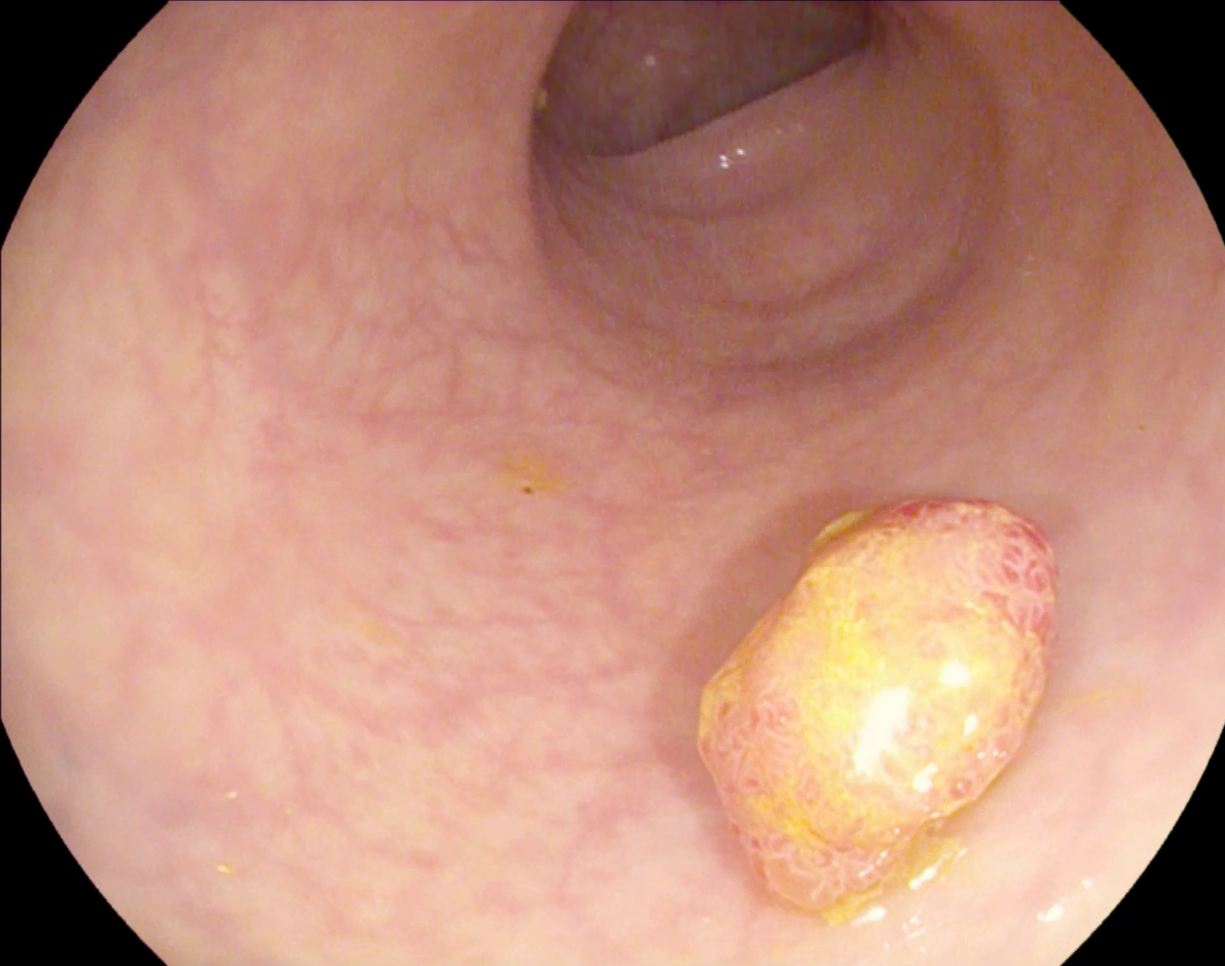} & \includegraphics{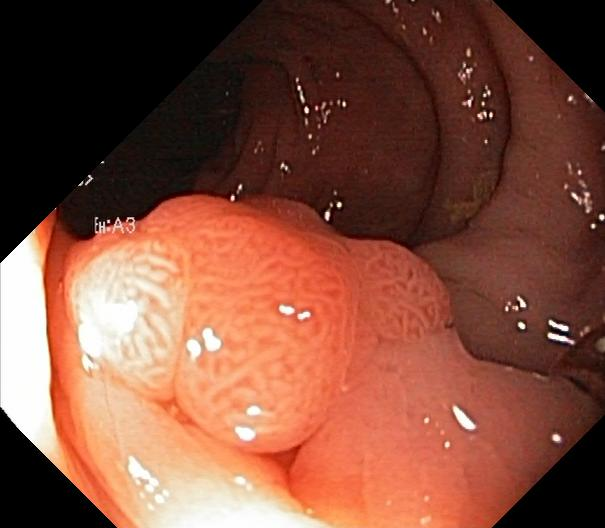}\\

    \raisebox{5.0\height}{\boldblue{$f^l$}}& 
    \includegraphics{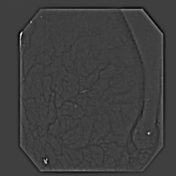} & 
    \includegraphics{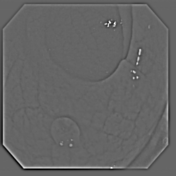} &
    \includegraphics{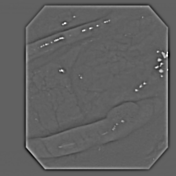} & 
    \includegraphics{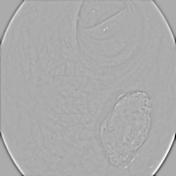} & 
    \includegraphics{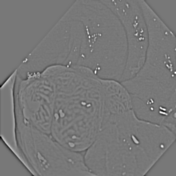} \\

    \raisebox{5.0\height}{\boldred{GT}}& 
    \includegraphics{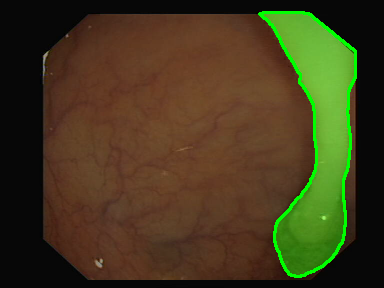} &
    \includegraphics{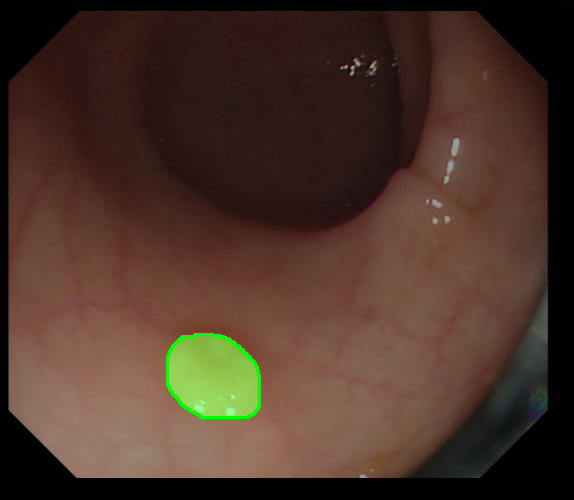} & 
    \includegraphics{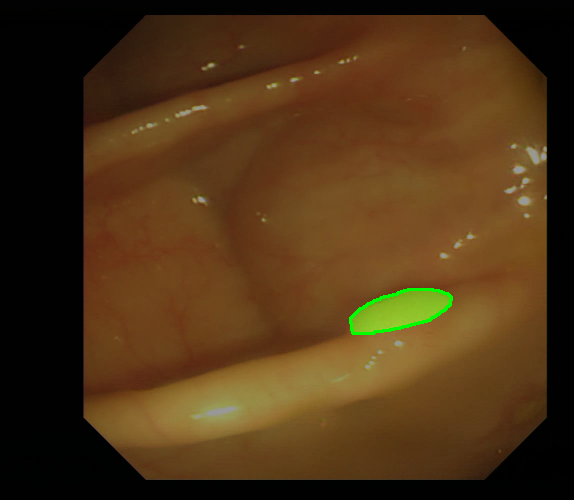} &
    \includegraphics{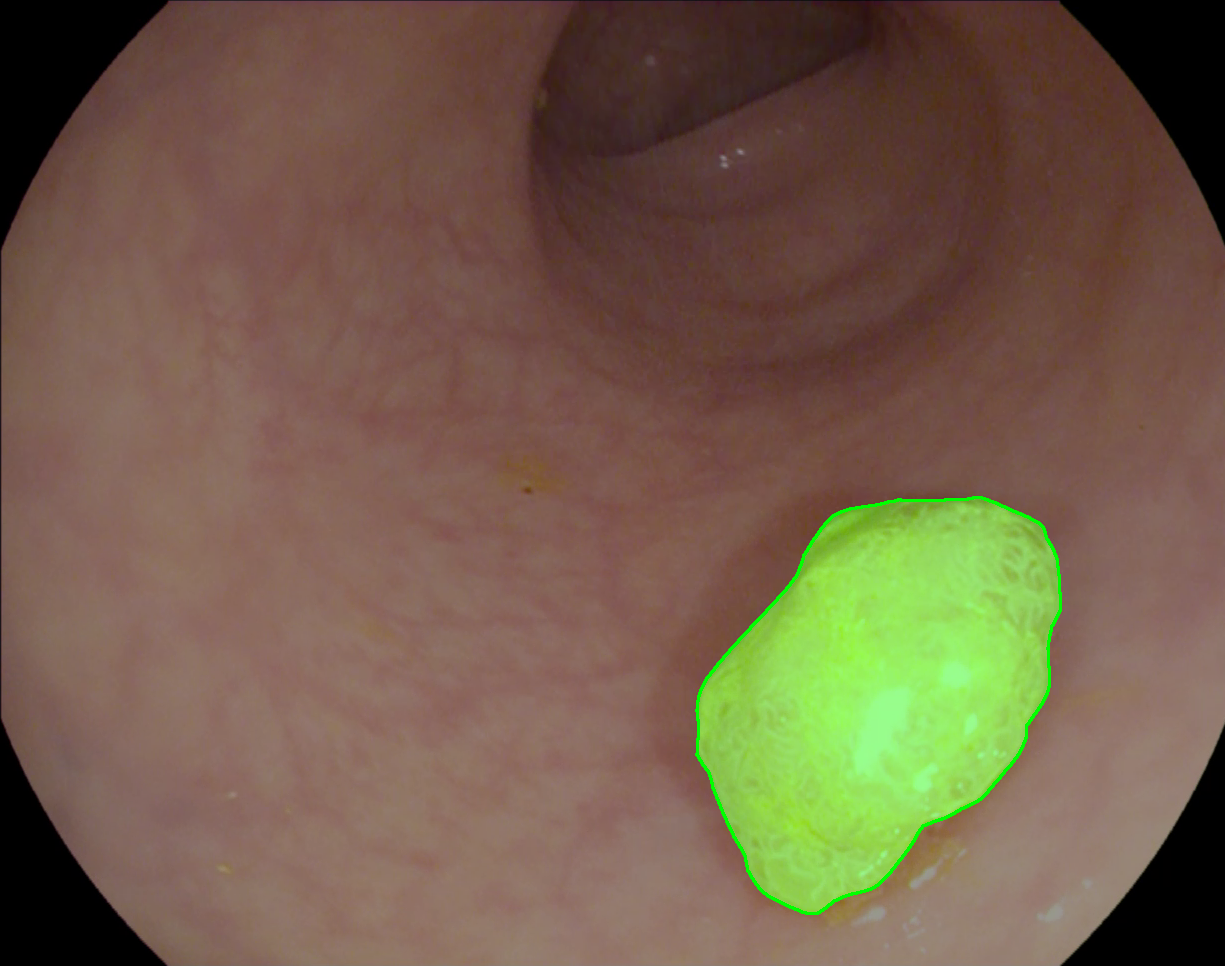} & 
    \includegraphics{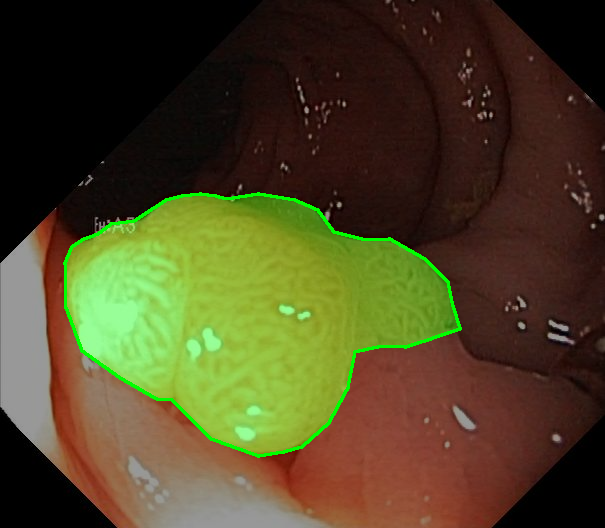}\\
   
\end{tabularx}
\caption{Visualization of the level-1 Laplacian pyramid in polyp images, which is basically the same with the high-frequency feature $f^l$. From left to right: the images come from the ClinicDB, ColonDB, CVC-300, ETIS and Kvasir datasets, respectively.}\label{fig:laplacian}
\end{center}
\end{figure*}

\begin{figure*}[!h]
\setkeys{Gin}{width=\linewidth,height=\linewidth}
\renewcommand{\arraystretch}{0.3}
\setlength{\tabcolsep}{1pt}
\begin{center}
\begin{tabularx}{0.8\textwidth}{sXXXX}  
    
    \raisebox{5.0\height}{Image}& 
    \includegraphics{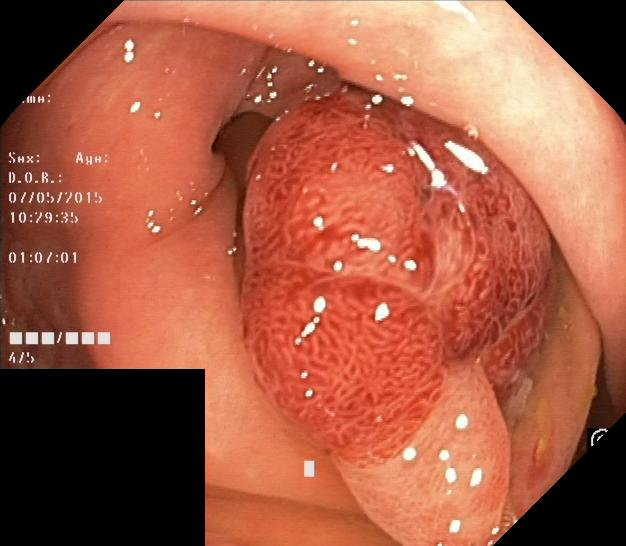} &
    \includegraphics{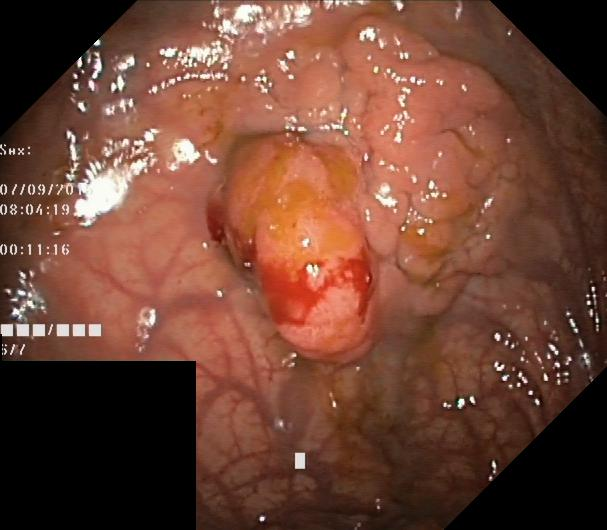} &
    \includegraphics{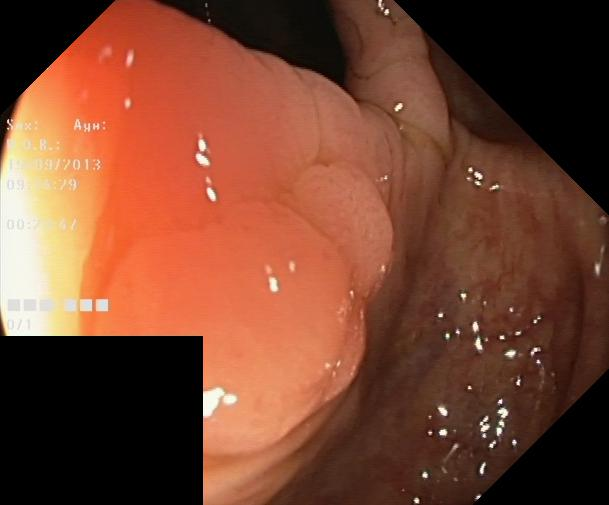}&
    \includegraphics{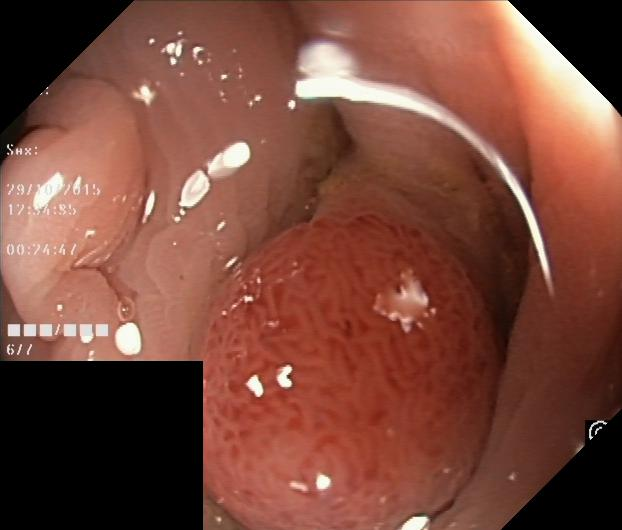}\\

    \raisebox{5.0\height}{\boldblue{$f^l$}}& 
    \includegraphics{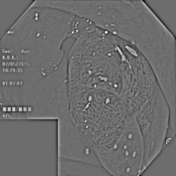}&
    \includegraphics{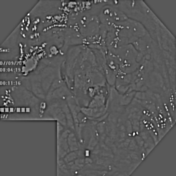} &
    \includegraphics{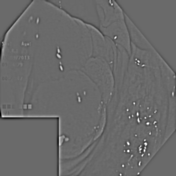}&
    \includegraphics{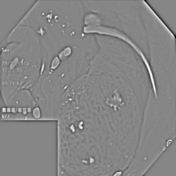}\\

    \raisebox{5.0\height}{\boldred{GT}}& 
    \includegraphics{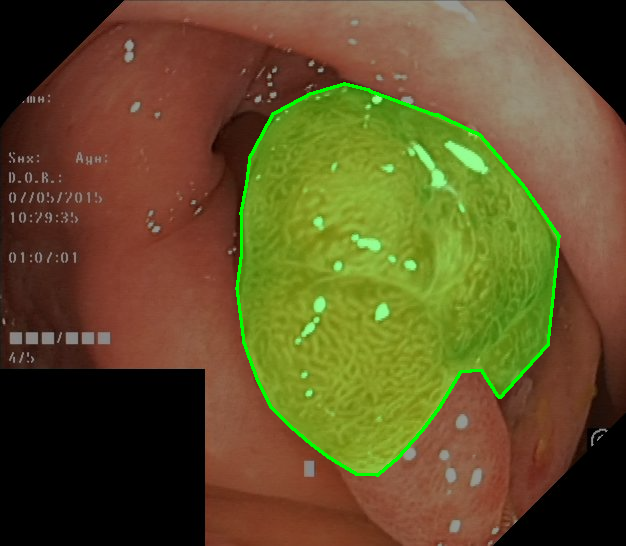} &
    \includegraphics{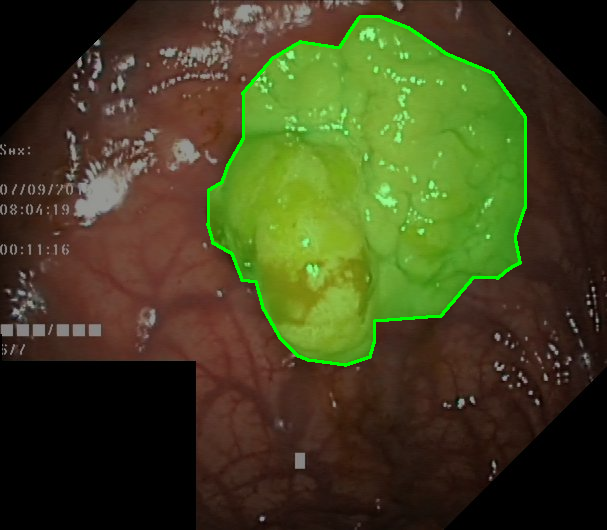}&
    \includegraphics{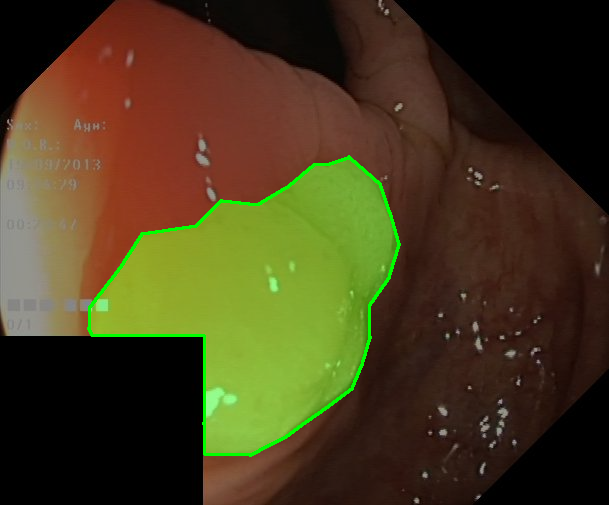}&
    \includegraphics{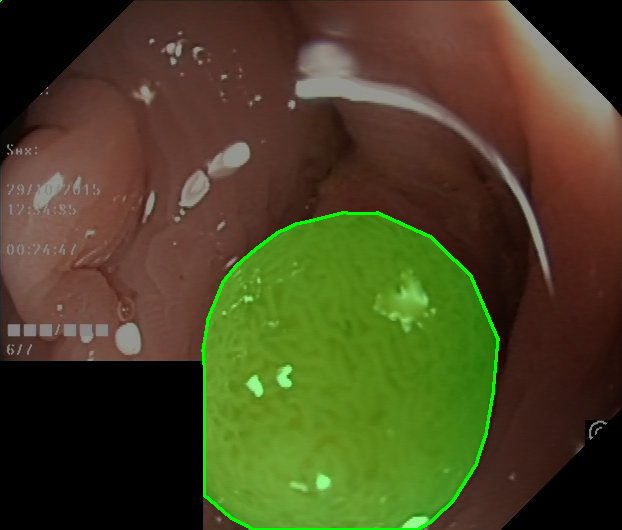}\\
   
\end{tabularx}
\caption{Visualization of the high-frequency feature $f^l$ in Kvasir images.}\label{fig:kvasir}
\end{center}
\end{figure*}

As previously mentioned, our Edge-Guided Attention (EGA) module takes three inputs, with one of them being the high-frequency feature denoted as $f^l$. In most cases, $f^l$ corresponds to $L_1(I)$, where $L_1$ represents the first-level image of the Laplacian pyramid, and $I$ is the input image. The visual representation of $f^l$ can be seen in Figure \ref{fig:laplacian}.

Obviously, the boundaries between different elements in the image are highlighted. The nuances of the texture are also exposed, highlighting its ability to preserve the high-frequency information inherent in the Laplacian pyramid. Leveraging this capability, the high-frequency feature $f^l$ effectively guides the model's attention towards edges relevant to polyps, thus enhancing predictions at each decoder level.

Upon examining polyp images from the Kvasir dataset, we've come to realize that the mucous membrane (background) of this dataset is considerably intricate. Consider the final image in Figure \ref{fig:laplacian} and Figure \ref{fig:kvasir} as examples; distinguishing between the appearance of polyps and the surrounding colonic mucosa proves to be quite challenging, even for human observers. This complexity is why the high-frequency feature $f^l$ extracted from the Kvasir dataset might contain noise, as it encompasses a significant amount of edge information from the mucous membrane. Consequently, this noise contributes to the scenario where our method, without the incorporation of $f^l$, achieves the highest score within the Kvasir dataset.

\section{Visualization of the decoded predicted feature}
\begin{figure*}[!h]
\centering
\includegraphics[width=\textwidth]{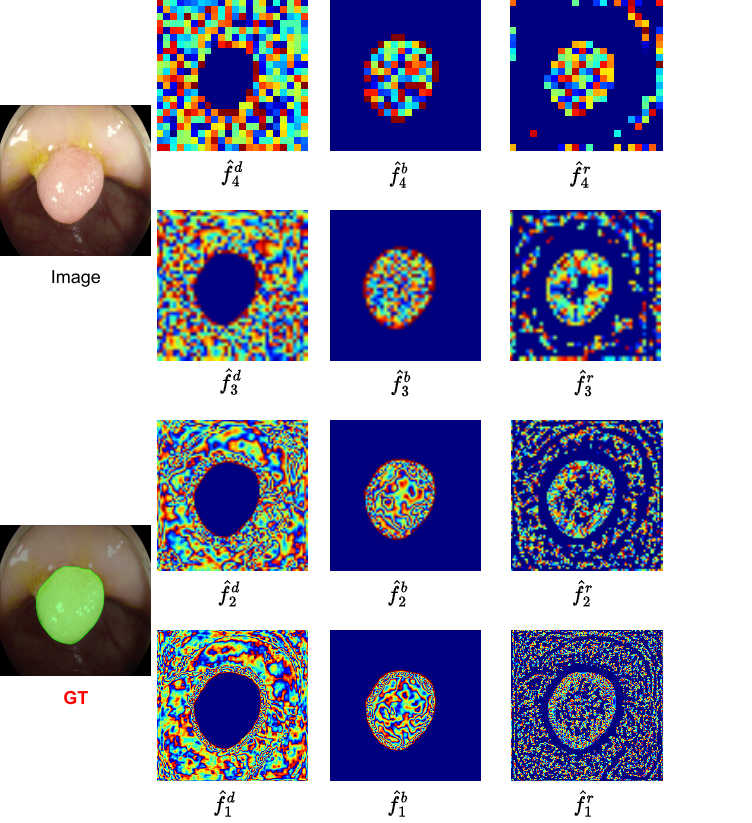}
\caption{Heatmap visualization of the decoded predicted feature $\hat{f}^d_{i}$, the boundary attention map $\hat{f}^b_{i}$ and the reverse attention map $\hat{f}^r_{i}$.}\label{attentionprediction}

\end{figure*}

Another input of our Edge-Guided Attention (EGA) module at certain layer $i^{th}$ is the predicted feature map at a higher layer $(i+1)^{th}$, denoted as $\hat{f}^d_{i+1}$.
The $\hat{f}^d_{i+1}$ is then decomposed into the reverse attention map $\hat{f}^r_{i+1}$ and the boundary attention map $\hat{f}^b_{i+1}$.
Figure \ref{attentionprediction} depicts the heatmap of an example for $\hat{f}^d_{i}$, 
$\hat{f}^b_{i}$ and $\hat{f}^r_{i}$ during the decoding phase where $i = [1, 2, 3, 4]$. As we can see, the boundary attention $\hat{f}^b_{i}$, which is attained from the Laplacian operator, efficiently concentrates on the boundary between the polyp object and the surrounding mucous membrane. This feature map can force the lower decoder layer to focus on the boundary of the prediction map from the higher layer better. The reverse attention $\hat{f}^r_{i}$, also shown in Figure \ref{attentionprediction}, guides the model to separate the polyp and mucous membranes by removing and refining the imprecise prediction map from the higher layer. By combining $\hat{f}^b_{i}$ and $\hat{f}^r_{i}$, the model can have the ability to recognize the whole object regions and better distinguish polyp from the background by accurate boundary.

